\documentclass[letterpaper]{article} % DO NOT CHANGE THIS
\usepackage{aaai24}  % DO NOT CHANGE THIS
\usepackage{times}  % DO NOT CHANGE THIS
\usepackage{helvet}  % DO NOT CHANGE THIS
\usepackage{courier}  % DO NOT CHANGE THIS
\usepackage[hyphens]{url}  % DO NOT CHANGE THIS
\usepackage{graphicx} % DO NOT CHANGE THIS
\usepackage{natbib}  % DO NOT CHANGE THIS AND DO NOT ADD ANY OPTIONS TO IT
\usepackage{caption} % DO NOT CHANGE THIS AND DO NOT ADD ANY OPTIONS TO IT
\usepackage[commentColor=blue]{algpseudocodex}
\usepackage{algorithm}

\usepackage{newfloat}
\usepackage{listings}
\DeclareCaptionStyle{ruled}{labelfont=normalfont,labelsep=colon,strut=off} % DO NOT CHANGE THIS
\floatstyle{ruled}
\newfloat{listing}{tb}{lst}{}
\floatname{listing}{Listing}

\usepackage{multirow}
\usepackage{amssymb}
\usepackage{amsmath}
\usepackage{booktabs}
\usepackage{subcaption}
\usepackage{textcomp,gensymb}
\usepackage{adjustbox}

\title{Catch-Up Mix: Catch-Up Class for Struggling Filters in CNN}
\author{
    Minsoo Kang\equalcontrib\textsuperscript{\rm 1,\rm 2}, Minkoo Kang\equalcontrib\textsuperscript{\rm 1}, Suhyun Kim\thanks{Corresponding Author.}\textsuperscript{\rm 1}
}
\affiliations{
    \textsuperscript{\rm 1}Korea Institute of Science and Technology, Republic of Korea\\
    \textsuperscript{\rm 2}Korea University, Republic of Korea\\
    \{minsoo.kang0918, mit240083, dr.suhyun.kim\}@gmail.com
}

\usepackage{bibentry}
\begin{document}

\maketitle
\begin{abstract}
Deep learning has made significant advances in computer vision, particularly in image classification tasks. Despite their high accuracy on training data, deep learning models often face challenges related to complexity and overfitting. One notable concern is that the model often relies heavily on a limited subset of filters for making predictions. This dependency can result in compromised generalization and an increased vulnerability to minor variations. While regularization techniques like weight decay, dropout, and data augmentation are commonly used to address this issue, they may not directly tackle the reliance on specific filters. Our observations reveal that the heavy reliance problem gets severe when slow-learning filters are deprived of learning opportunities due to fast-learning filters. Drawing inspiration from image augmentation research that combats over-reliance on specific image regions by removing and replacing parts of images, our idea is to mitigate the problem of over-reliance on strong filters by substituting highly activated features. To this end, we present a novel method called Catch-up Mix, which provides learning opportunities to a wide range of filters during training, focusing on filters that may lag behind. By mixing activation maps with relatively lower norms, Catch-up Mix promotes the development of more diverse representations and reduces reliance on a small subset of filters. Experimental results demonstrate the superiority of our method in various vision classification datasets, providing enhanced robustness.

\end{abstract}
% We take inspiration from image augmentation researches that address over-reliance on specific image regions through the removal and replacement of image parts.
% Our idea is to mitigate a problem of over-reliance on strong filters by replacing highly activated features.

\section{Introduction}
There have been remarkable improvements in deep learning algorithms, especially in computer vision fields. 
Since the ImageNet challenge~\cite{krizhevsky2012imagenet}, modern image classifiers have achieved near-human level accuracy in the image recognition task.
Although models may achieve high accuracy, they often face challenges related to their complexity and overfitting. Even slight variations or perturbations in input images, such as rotation, rescaling~\cite{Goodfellow-et-al-2016}, corruption~\cite{hendrycks2019benchmarking}, or adversarial attacks~\cite{szegedy2013intriguing,goodfellow2014explaining}, can lead to unexpected model behavior. 
For instance, a classifier trained on clean images from a sunny day may underperform with inputs from rainy conditions. 
Such vulnerabilities underscore the importance of robustness, especially when models tackle real-world scenarios subject to distributional shifts~\cite{hendrycks2019benchmarking}. 

There are several factors that can contribute to a model's limited generalization and robustness.
One particular phenomenon we focus on is the convolutional neural network (CNN) model's heavy reliance on a small subset of convolutional filters for making predictions.
Models relying on a limited number of filters often exhibit poor performance on test data and lack robustness against slight variations.
Furthermore, a concerning aspect is that rapidly trained weights can potentially make inaccurate judgments due to their tendency to learn dataset-specific biases~\cite{nam2020learning}. 
These biases may capture object-irrelevant patterns (e.g., background) rather than focusing on the object of interest.

To address this issue, regularization methods such as weight decay, dropout, and data augmentation are widely employed.
Dropout~\cite{srivastava2014dropout} tempers the model's tendency to over-rely on particular units by randomly deactivating them. 
Weight decay~\cite{ng2004feature}, known as $\ell_{2}$ regularization, encourages the model to have smaller weights by adding a penalty for large weights.
Data augmentation~\cite{devries2017improved_cutout, hinton2012improving, simonyan2014very, kang2023guidedmixup} encourages learning more robust and generalized representations by generating training data variants through assorted transformations.
However, those regularization methods do not directly address the model's reliance on specific filters.

\begin{figure*}[t]
\centering
\includegraphics[width=0.97\textwidth]{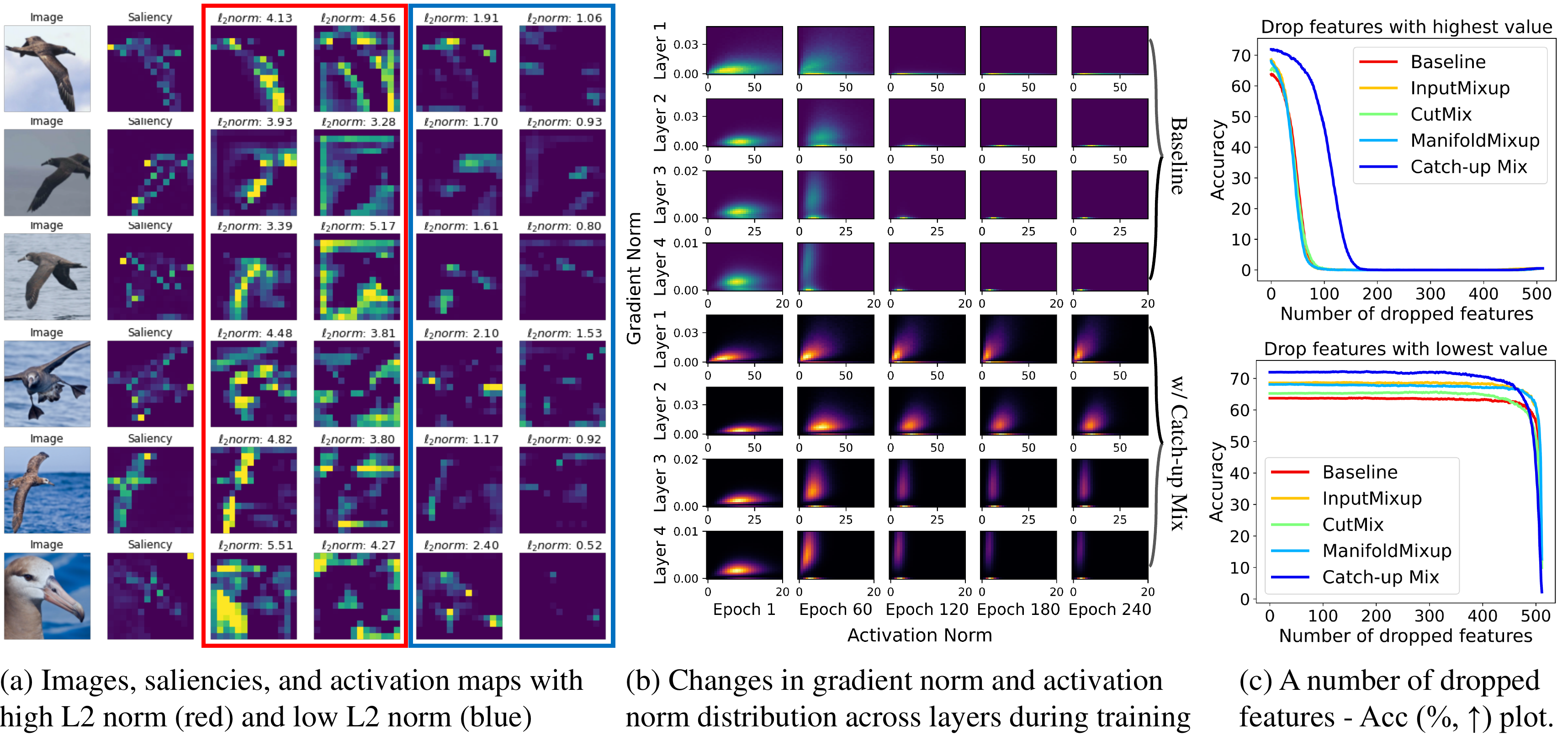}
\caption{
We visualize activation maps, $\ell_{2}$ norm, and gradients to understand how filters behave during training. (a) shows visualizations of images, saliency maps, and activation maps.
Activation maps are obtained from the third layer block output of ResNet-18 while training CUB-200 at 120 epochs. (b) represents how distributions of activation norm (x-axis) and gradient norm (y-axis) change over epochs. 
For example, among rows of `Baseline,' the gradient scale decreases significantly after 60 epochs, making it difficult to update the weight properly for the remaining process.
(c) depicts the model's accuracy as latent vectors are sequentially dropped by their value. This highlights that our method prompts the model to use diverse features.
}
\label{fig:activation} 
\end{figure*}

Our observations indicate that the heavy reliance on a small subset of filters in CNN can occur when slow-learning filters lack adequate opportunities to learn, especially when a small number of the fast-learning filters are sufficient to classify the training dataset accurately.
It is supported by the visualization of activation maps (Figure~\ref{fig:activation}a) and histograms, which illustrate the relation between the $\ell_{2}$ norms of activation maps and corresponding gradients (Figure~\ref{fig:activation}b).
Figure~\ref{fig:activation}a provides insights into filters' learning progress and impact during training.
We notice that fast-learning filters tend to exhibit higher $\ell_{2}$ norms of activations, often capturing biases like the background. 
On the other hand, lower-magnitude filters struggle to extract meaningful information from the input.

The problem is that slow-learning filters not only lag behind in their learning progress but are also deprived of future learning opportunities in the remaining epochs.
The main reason is that the current model already reaches a top-1 training accuracy of 95.81\% at 120 epochs.    
Thus, despite many filters failing to extract features properly at 120 epochs (Figure \ref{fig:activation}a), only small gradients will be generated, indicating minimal updates to these filters (Figure \ref{fig:activation}b).
This imbalance causes poor generalization performance, as evidenced by the top-1 validation accuracy rates of 48.21\%, 50.00\%, and 51.50\% achieved at 60, 120, and 150 epochs, respectively.
As a result, unfortunately, a significant portion of the training process becomes unproductive for these slow learners, hindering the model from acquiring generalized representations.

Throughout this analysis, we observe that slow-learning filters tend to lag behind during the training and have limited opportunities for the model to improve for the rest of the training process. 
Our design is inspired by image augmentation techniques which train models on defective images by removing (CutOut~\cite{devries2017improved_cutout}) or replacing (CutMix~\cite{yun2019cutmix}) parts of the image to learn different aspects of the data without relying on any particular region. 
Transposing this philosophy to the feature space, we intentionally exclude and supplant well-trained filters from a stochastic layer.
By training on these defective feature maps, we encourage the model to learn different features of data utilizing the `\textit{struggling filters}' without relying solely on a small number of strong filters. 
We denote this concept as a `\textit{catch-up class},' allowing slow-learners to be trained and contribute to the model's decision-making process.

In this paper, we present Catch-up Mix, a novel feature-level mixup method designed to offer learning opportunities to a wide range of filters during training, focusing on filters that may fall behind.
Catch-up Mix ensures broader training exposure for less-evolved filters by mixing activation maps of pairs with relatively low $\ell_{2}$ norms and updating gradients by excluding filters with high magnitudes.
Note that choosing a `relative' smaller norm doesn't mean that every selected filter is under-trained; instead, it provides a variety of filter combinations to mitigate over-reliance issues.
Consequently, the trained model makes robust predictions without relying heavily on a small subset of filters.

The experiments show that our method outperforms other methods on various vision classification datasets. 
Furthermore, we designed Catch-up Mix to utilize a greater number of filters to identify and capture the characteristics and patterns in the input data, leading us to expect high levels of robustness and feature diversity.
To validate this, we assessed its robustness through various means, such as adversarial attacks, deformation, and data corruption. 
Since our main concern, the problem of struggling filters, is intrinsically tied to the diversity of the training dataset and the capacity of the model, we assess outcomes both with limited datasets and large-scale datasets.
Moreover, the results from out-of-distribution detection and loss landscape visualization further emphasize its robustness and generalization capabilities.

\begin{figure*}[th]
\centering
\includegraphics[width=0.85\textwidth]{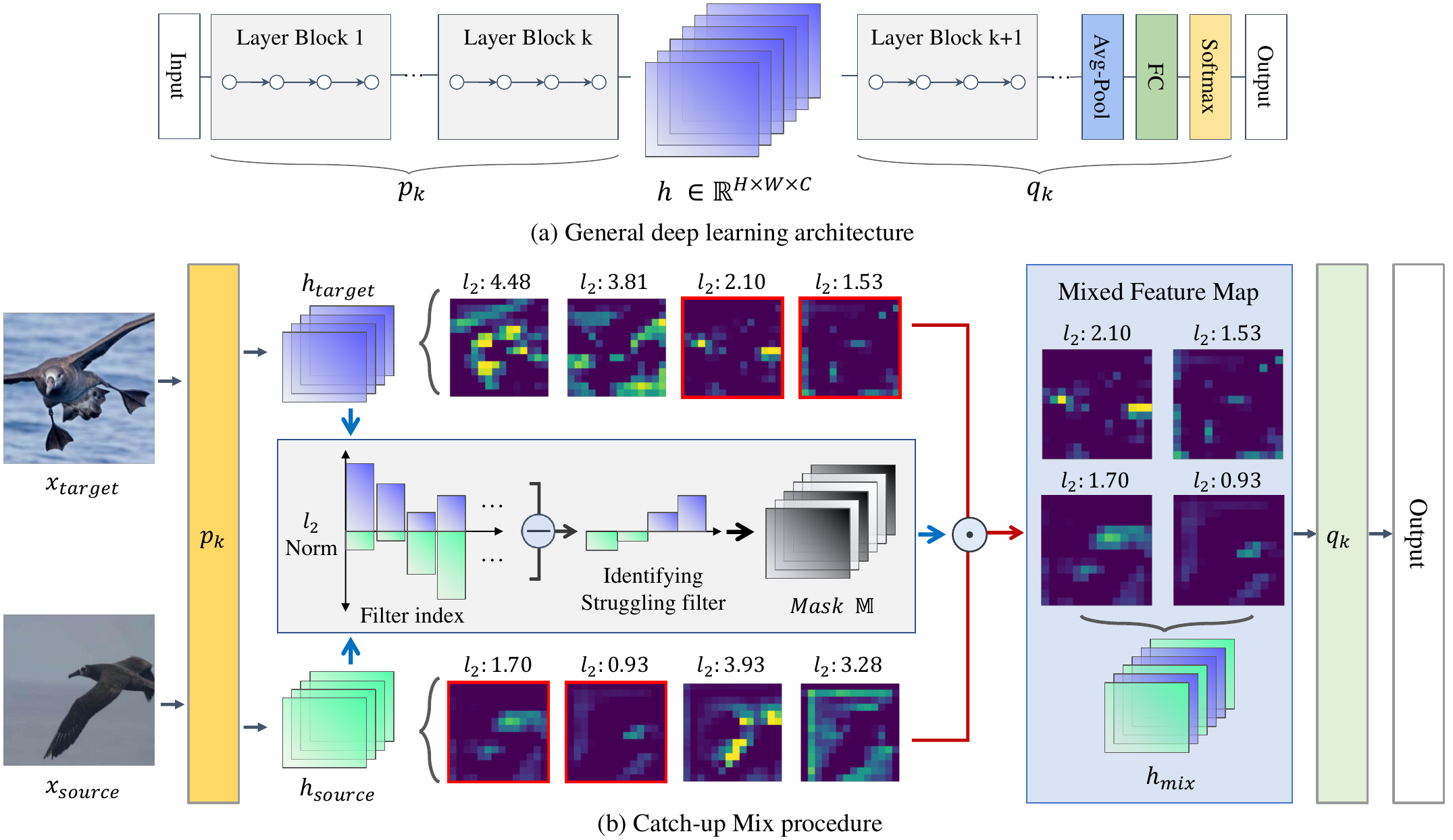}
\caption{
The overall framework and augmentation process of Catch-up Mix. (a) represents a general deep learning architecture, and (b) outlines the procedure of Catch-up Mix.
First, we compare the magnitudes of activation maps using their $\ell_{2}$ norms.
A mask $\mathbb{M}$ is generated to mix the activation maps from the selected pair with relatively low $\ell_{2}$ norms. 
}
\label{fig:main}
\end{figure*}

\section{Method}
\subsection{Preliminaries}
Mixup augmentation is performed on a pair of input images or feature maps of the network to enhance generalization capabilities. A model with input-level mixup can be expressed as $f(mix_{\lambda}(x,x'))$, where $x,x' \in X$ represents paired images from the input batch. $f : X \rightarrow \mathbb{R}^{N}$ is a given network where $N$ is the number of classes, and $\lambda$ is a mixing ratio. Usually, $\lambda$ is randomly sampled from $Beta(\alpha,\alpha)$ distribution, where $\alpha$ is a hyper-parameter.
For example, InputMixup~\cite{zhang2018mixup} can be expressed as follows:
\begin{equation}
mix_{\lambda}(x,x') = \lambda \cdot x + (1-\lambda) \cdot x'.
\end{equation}
To formalize the network with feature-level mixup, we define two functions as illustrated in Figure~\ref{fig:main}: $p_{k} : X \rightarrow \mathbb{R}^{C \times H \times W}$ and $q_{k} : \mathbb{R}^{C \times H \times W} \rightarrow \mathbb{R}^{N}$, that are mapping from input image to the feature map of layer $k$ and from feature map of layer $k$ to model output.
Throughout this paper, we denote $h_{k} \in \mathbb{R}^{C \times H \times W}$ as the feature map of layer $k$ and $h_{k}^{i} \in \mathbb{R}^{H \times W}$ as the activation map of $i$th filter in layer $k$.
Thus, $p_{k}$ and $q_{k}$ satisfy $f(x) = q_{k}(p_{k}(x))$ and $h_{k}=p_{k}(x)$. Those mixup methods $f_{\lambda}(x,x')$ can be expressed as follows:
\begin{equation}
    \left\{
    \begin{matrix}
    f(mix_{\lambda}(x,x')),&\text{Input-level mixup,}\\ 
    q_{k}(mix_{\lambda}(p_{k}(x),p_{k}(x'))),&\text{Feature-level mixup.}
    \end{matrix}
    \right.
\end{equation}
In the image classification task, each image sample $x \in X$ has a one-hot encoded label vector $y \in \{0,1\}^{N}$ where $N$ is the number of classes.
Labels should also be interpolated when a mixup method is used. The following equation is the mixed label determined by ground truth labels weighted by the mixing ratio $\lambda$:
\begin{equation}
    y_{mix} = \lambda \cdot y + (1-\lambda) \cdot y'.
\end{equation}

\subsection{Proposed Method}
In this section, we introduce Catch-up Mix, a method designed to address the over-reliance on small subsets of filters and enhance the model's generalization capability.
Our observation is that activations with low $\ell_{2}$ norms tend to be relatively underdeveloped during the training process (Figure~\ref{fig:activation}a).
Also, if a model with less developed filters can still achieve high training accuracy, there will be a scarcity of gradients (Figure~\ref{fig:activation}b).
This indicates that less-developed filters may not have sufficient learning opportunities in the subsequent epochs and not reach their full potential.

Based on this observation, our approach randomly selects a layer and mixes activation maps that exhibit relatively less development for a given input.
By making uncertain predictions with low confidence based on these mixed features, we can update the corresponding filters that do not capture object characteristics well, thereby offering them more substantial training opportunities.
Catch-up Mix enables the model to prioritize the improvement of less developed filters and mitigates its reliance on a few highly influential filters. 
As a result, Catch-up Mix enhances the model's generalization ability and reduces its vulnerability to overfitting.

For each iteration of Catch-up Mix, a random layer $k$ is selected, and the batch input is passed through the model up to that $k$-th layer to obtain the source and target feature maps, denoted as $h_k$ and $h_k'$.
For a given input, the influence of each filter on predictions is assessed by computing the $\ell_{2}$ norm of its activations $h_{k}^c$, denoted as Filter Influence ($FI$),
\begin{equation}
    FI = \{FI_{c} | FI_{c} = ||h_{k}^c||_{2},\quad \text{for all } c \in \mathcal{C} \}.
\end{equation}

Next, we compare the sum-to-one normalized $FI$s of source and target mixup pairs, and the activation maps with the relatively lower filter influence (Relative Filter Influence; $RFI$) are selected.
This normalization is necessary because the scale of the activation map, and thus the scale of the $FI$s, can vary significantly depending on the input. 
If $RFI_i$ is high, $i$-th filter has a stronger influence on predicting the source image compared to the target image, and vice versa.
\begin{equation}
    \begin{aligned}
        RFI =\quad &\{RFI_c | RFI_c = \frac{FI_c}{\sum_{c=0}^{\mathcal{C}}(FI_c)} - \frac{FI'_c}{\sum_{c=0}^{\mathcal{C}}(FI'_{c})}, \\
         &\text{for all } c\in \mathcal{C} \}.
    \end{aligned}
\end{equation}

Using $RFI$ and mixing ratio $\lambda$, we generate a binary mixup mask that determines which filters' activation map should be propagated to the next layer. 
Given the number of filters in the layer and $\lambda$, we calculate the number of filters to retain from source feature map $N_{mix} = \lfloor{\lambda \times |\mathcal{C}|}\rfloor$.
Next, we generate filter-wise activation mask $\mathbb{M}$ that drops the activation maps with top $|\mathcal{C}| - N_{mix}$ $RFI$ values.
The sum of the mask elements equals $N_{mix}$.
Finally, the mixed feature map is computed as follows:
\begin{equation}
h_{mix} = \mathbb{M} \odot h + (\mathbb{I} - \mathbb{M}) \odot h',   
\end{equation} 
where $\mathbb{I} \in \{1\}^C$ is a binary mask filled with ones, $\odot$ is a filter-wise multiplication operation.
The output of network $f_{\lambda}$ with Catch-up Mix at layer $k$ can be expressed as follows:
\begin{equation}
f_{\lambda}(x,x') = q_{k}( \mathbb{M} \odot p_{k}(x) + (\mathbb{I} - \mathbb{M}) \odot p_{k}(x')).
\end{equation}

For better understanding, we explain how Catch-up Mix operates using ResNet~\cite{he2016deep}, which consists of five-layer blocks: the first convolutional layer as the first layer block and four stages of ResNet as the other four layer blocks. 
Here, we uniformly sample the mixup layer $k$ from the set $K=\{0,1,2,3,4,5\}$ every iteration, where layer 0 corresponds to the input layer. 
If $k=0$, CutMix~\cite{yun2019cutmix} is applied, and if $k>0$, Catch-up Mix is applied to feature maps of $k$th layer block.
Layer blocks for applying Catch-up Mix can be set arbitrarily.
For models with many layers or multi-stage, we can use Catch-up Mix, applying an appropriate mixup layer set.
After, we assign a new label for the mixed feature map according to the mixing ratio $\lambda$.

\section{Experiments}

In this section, we extensively evaluate the performance and robustness of Catch-up Mix on various datasets. 
First, we compare the generalization performance of the model trained with proposed methods on general classification datasets: CIFAR-100~\cite{Krizhevsky2009learning}, and Tiny-ImageNet~\cite{chrabaszcz2017downsampled}.
In addition, we measure the augmentation overhead of mixup methods.
Next, we evaluate the robustness of classifiers against adversarial attacks, data corruption, and deformation to validate the improvement achieved by Catch-up Mix. 

Furthermore, we validate that our performance improvements are not limited to specific datasets and architectures.
We also evaluate our methods on standard fine-grained datasets: CUB-200-2011 (CUB)~\cite{WahCUB_200_2011}, Stanford-Cars (Cars)~\cite{cars}, and FGVC-Aircraft (Aircraft)~\cite{aircraft}.
Note that achieving high performance on fine-grained datasets requires the model to have the capability of capturing the fine details in images.
Since the `\textit{struggling filter}' problem we want to address is related to the volume of the dataset, we compare and analyze the performance of training on small datasets and a large dataset, ImageNet~\cite{krizhevsky2012imagenet}. Also, we evaluate the out-of-distribution detection performance using ImageNet-O~\cite{hendrycks2021natural} to assess the generalization performance.

Throughout this paper, we present the results of experiments aimed at evaluating the effectiveness of Catch-up Mix across various architectures: ResNet~\cite{he2016deep}, PreActResNet~\cite{he2016identity}, DenseNet~\cite{huang2017densely}, PyramidNet~\cite{han2017deep}, VGGNet~\cite{simonyan2014very}, MobileNetv2~\cite{sandler2018mobilenetv2}, Wide-ResNet~\cite{zagoruyko2016wide}, and ResNeXt~\cite{xie2017aggregated}.
Lastly, we compare the loss landscape to visualize the regularization effect of Catch-up Mix.

\begin{table}[t]
\centering
\small{
\begin{tabular}{lccccccc}
\toprule 
 \multirow{2}*{\textbf{Method}} & \multicolumn{1}{c}{Top-1} & \multicolumn{1}{c}{FGSM} & \multicolumn{1}{c}{Time/epoch} & \multicolumn{1}{c}{mCE} \\
 & \multicolumn{1}{c}{Err (\%)} & \multicolumn{1}{c}{Err ($\%$)} & \multicolumn{1}{c}{(s, $\downarrow$)} & \multicolumn{1}{c}{(\%)} \\
\midrule
Baseline    & 22.98 & 84.93 & 103.8$\pm$0.5 & 51.43                           \\
InputMixup  & 19.26 & 75.72 & 104.4$\pm$0.5 & 44.84                           \\
CutMix      & 19.95 & 73.89 & 104.5$\pm$0.7  & 53.74                           \\
PuzzleMix   & 18.55 & 79.33 & 275.8$\pm$2.5  & 46.21                           \\
SaliencyMix & 19.43 & 81.25 & -   & 47.85                           \\
Co-Mixup    & 20.06 & 87.94 & 894.0$\pm$6.5 & 53.72                           \\
\midrule
Manifold    & 19.32 & 81.96 & 107.1$\pm$0.4 & \underline{44.32}                     \\
MoEx        & 19.45 & 80.41 & 105.9$\pm$0.5 & 52.52                           \\
RecursiveMix& 19.42 & 89.58 & -             & 48.52                         \\
AlignMixup  & \underline{19.20}  & \underline{69.84} & 118.1$\pm$0.5 & 44.71   \\
\midrule
\textbf{Catch-up Mix} & \textbf{17.76} & \textbf{68.16} & 107.8$\pm$0.6 & \textbf{43.76}   \\ 
\bottomrule
\end{tabular}
}
\caption{
Top-1 / FGSM error rate (\%, $\downarrow$), and computational cost (s, $\downarrow$) of training PreActResNet-18 with mixup methods on CIFAR-100.
Also, we evaluate the mean Corruption Error (mCE, \%, $\downarrow$) rate on CIFAR-100-C using a pre-trained model on CIFAR-100 using corresponding mixup methods. 
} \label{tab:cifar_acc_corrupted}
\end{table}

\begin{table*}[t]
\centering
\small{
% \resizebox{\columnwidth}{!}{
\begin{tabular}{lccc|ccccccccc}
\toprule
\multirow{3}*{\textbf{Method}} & \multicolumn{3}{c}{General} & \multicolumn{8}{|c}{Deformation Performance Accuracy (\%, $\uparrow$)} \\ 
 \cmidrule(lr){2-4} \cmidrule(lr){5-12} 
 & \multicolumn{1}{c}{Top-1} & \multicolumn{1}{c}{FGSM} & \multicolumn{1}{c}{Time/epoch} 
 & \multicolumn{2}{|c}{Random Rotate} &\multicolumn{2}{|c}{Random Sheering} &\multicolumn{4}{|c}{Zoom In/Out} \\
& \multicolumn{1}{c}{Err (\%, $\downarrow$)} & \multicolumn{1}{c}{Err ($\%$, $\downarrow$)} & \multicolumn{1}{c}{(s, $\downarrow$)} &\multicolumn{1}{|c}{$\pm20\degree$} & \multicolumn{1}{c}{$\pm40\degree$} & \multicolumn{1}{|c}{$\pm28.6\degree$}  & $\pm57.2\degree$ &\multicolumn{1}{|c}{60\%}  & 80\% & 120\% & 140\% \\
\midrule
Baseline   & 36.98          & 89.34 & 21.2$\pm$0.1 & 53.96          & 42.99          & 54.14          & 37.62          & \underline{20.41} & \underline{47.51}          & 50.67           & 40.76           \\
InputMixup      & 35.32          & 91.02 & 21.3$\pm$0.2 & 53.20          & 42.45          & 54.94          & 39.44          & 17.88          & 47.14          & 52.54           & 43.43           \\
CutMix     & 34.58          & 87.00 & 21.4$\pm$0.2 & \underline{56.63} & \underline{45.13} & 57.70 & \underline{43.00} & 12.69 & 44.72 & 47.43 & 38.65 \\
PuzzleMix  & \underline{32.98}   & 87.22 & 64.4$\pm$0.8 & 53.93 & 40.90 & 56.02 & 40.69 & 9.26 & 38.94 & 47.10 & 36.43 \\
SaliencyMix &  34.76        & 93.75 & - & 54.78 & 44.14 & 56.75 & 41.80 & 15.18 & 44.85 & 49.48 & 41.45 \\
Co-Mixup   & 34.58          & 91.11 & 244.7$\pm$1.6 & 53.12 & 39.89 & 52.22 & 37.86 & 7.72 & 33.52 & 40.15 & 32.05 \\
\midrule
Manifold & 35.46          & 88.74 & 21.8$\pm$0.2 & 54.37          & 43.65          & 55.78          & 39.42          & 15.51          & 46.32          & 49.78           & 39.12           \\
MoEx       & 34.27          & 89.49 & 21.7$\pm$0.2 & 56.23          & 44.56  & \underline{57.79}   & 41.24          & 14.66          & 45.52          & 48.55           & 40.70           \\
RecursiveMix & 33.97        & 92.88 & - & 56.01          & 43.74          & 56.75          & 41.40          & 12.03          & 44.11          & 50.06           & \underline{44.09}           \\
AlignMixup & 34.03          & \underline{86.56} & 35.4$\pm$0.3 & 54.87          & 43.62          & 57.25          & 41.38          & 12.36          & 46.07          & \underline{53.55}           & 42.57           \\
\midrule
\textbf{Catch-up Mix}      & \textbf{30.73} & \textbf{84.22} & 22.2$\pm$0.2 & \textbf{60.20} & \textbf{47.87} & \textbf{61.52} & \textbf{44.06} & \textbf{24.03} & \textbf{55.46} & \textbf{58.34}  & \textbf{48.32}  \\
\bottomrule
\end{tabular}
}
\caption{
Top-1 / FGSM error rate (\%, $\downarrow$), and computational cost (s, $\downarrow$) of training the model with mixup methods on Tiny-ImageNet datasets using PreActResNet-18. Top-1 accuracy rates (\%, $\uparrow$) on the test datasets of Tiny-ImageNet with various deformations. We evaluate the PreActResNet-18 model pre-trained on the original Tiny-ImageNet dataset.
} \label{Tab:ti_acc_deformation_acc}
% }
\end{table*}

\subsection{Performance on General Classification Datasets} \label{exp:general}

We evaluate the performance of our method on two benchmark datasets: CIFAR-100 and Tiny-ImageNet. 
We use PreActResNet-18 as the backbone architecture and reproduce existing mixing data-based augmentation methods: InputMixup~\cite{zhang2018mixup}, CutMix~\cite{yun2019cutmix}, PuzzleMix~\cite{kim2020puzzle}, SaliencyMix~\cite{uddin2020saliencymix}, Co-Mixup~\cite{kim2020co}, SnapMix~\cite{huang2021snapmix}, ManifoldMixup (Manifold)~\cite{verma2019manifold}, MoEx~\cite{li2021feature_moex}, AlignMixup~\cite{venkataramanan2022alignmixup}, and RecursiveMix~\cite{yangrecursivemix}. 
We train each model for 1200 (Tiny-ImageNet) and 2000 epochs (CIFAR-100) using hyperparameter settings from AlignMixup. 
For method-specific mixup parameters, we follow the paper and the official code. 
As shown in Table \ref{tab:cifar_acc_corrupted}, Catch-up Mix achieves the best generalization performance compared to other mixup baselines. 
On Tiny-ImageNet, Catch-up Mix outperforms PuzzleMix by about 2\%, and on CIFAR-100, it outperforms AlignMixup by about 1.5\%. 
PuzzleMix and AlignMixup were the second-best performing methods on each dataset. 

\subsubsection{Training Overhead}
Also, we measure the training overhead of mixup methods by training the classifier for 20 epochs on CIFAR-100 with RTX2080 and Tiny-ImageNet with TITAN RTX. 
For a fair comparison, we measured the time without any other workloads running to avoid interference, and we did not use multiprocessing support to accelerate mixup augmentation. 
As Catch-up Mix does not require additional forward and backward propagation, the results show that it has only negligible overhead.

\subsubsection{Robustness against Adversarial Attacks}
In this section, we assess the adversarial robustness of classifiers trained with various mixup methods.
To evaluate adversarial robustness, we measure the model's performance on adversarial examples generated by applying Fast Gradient Sign Method (FGSM)~\cite{goodfellow2014explaining} to the test dataset with a 4/255 $l_{\infty}$ epsilon ball.
The results show that Catch-up Mix improves the FGSM Top-1 error rate over the best-performing baseline by 2.34\% on Tiny-ImageNet and 1.68\% on CIFAR-100, as shown in Table~\ref{tab:cifar_acc_corrupted} and~\ref{Tab:ti_acc_deformation_acc}.

\subsubsection{Robustness against Data Corruption}
Also, We evaluate the robustness against data corruption using CIFAR-100-C, synthesized by corrupting the CIFAR-100 test dataset with 19 types of corruptions, such as fog, snow, brightness, blur, and noise, at five different strengths. The mean Corruption Error (mCE, \%, $\downarrow$), calculated across all types of corruption, is reported in Table~\ref{tab:cifar_acc_corrupted}. 
Our results demonstrate that Catch-up Mix achieves the highest robustness against data corruption, even without employing additional training processes tailored explicitly for data corruption.

\subsubsection{Robustness against Deformation}
In real-world scenarios, objects within images can be situated at varying distances or orientations, leading to irregularly rotated or zoomed-in/out instances. 
These deformations necessitate the development of a model with robust generalization capabilities to accommodate various forms of deformation. 
Following ManifoldMixup~\cite{verma2019manifold}, we evaluate a pre-trained classifier (refer to Table~\ref{Tab:ti_acc_deformation_acc}) on Tiny-ImageNet datasets subjected to random rotation, random sheering, and zoom in/out to assess deformation robustness.
As shown in Table~\ref{Tab:ti_acc_deformation_acc}, the model pre-trained with Catch-up Mix exhibits high robustness against deformations, successfully classifying test instances across all deformation variations compared to all the other mixup methods.

Ensuring robustness is a critical aspect of reliable deep learning models.
It is vital to achieve robustness solely through a training method rather than training with corrupted samples that may limit generalization to specific distortions.
For example, when a model is trained exclusively with distorted data, it specializes in recognizing the specific corruptions encountered during training, thereby hindering its ability to generalize to unseen distortions~\cite{geirhos2018generalisation}.
As our empirical results show, Catch-up Mix, which emphasizes balanced filter utilization, successfully improves robustness without additional effort.

\begin{table}[t]
\centering
\resizebox{0.98\columnwidth}{!}{%
\small{
\begin{tabular}{lcccccc}
\toprule
 \multirow{2}{*}{\textbf{Method}} & \multicolumn{2}{c}{\textbf{CUB}} & \multicolumn{2}{c}{\textbf{Cars}} & \multicolumn{2}{c}{\textbf{Aircraft}} \\
 \cmidrule(lr){2-3} \cmidrule(lr){4-5} \cmidrule(lr){6-7} &
  \multicolumn{1}{c}{R18} &  \multicolumn{1}{c}{D121} &
  \multicolumn{1}{c}{R18} &  \multicolumn{1}{c}{D121} &
  \multicolumn{1}{c}{R18} &  \multicolumn{1}{c}{D121} \\
\midrule
Baseline     & 63.77 & 68.88 & 82.37  & 85.44 & 77.92  & 80.01 \\
InputMixup   & 67.24 & 74.28 & 85.67  & 88.58 & 80.38  & 81.24 \\
CutMix       & 64.08 & 74.62 & 86.57  & 89.10 & 80.05  & 81.15 \\
PuzzleMix    & 69.33 & 76.32 & 86.51  & 89.96 & 79.18  & 82.59 \\
SaliencyMix  & 61.74 & 67.53 & 85.66  & 88.90  & 79.78  & 80.88 \\
Co-Mixup     & \underline{71.82} & \underline{76.88} & 87.31 & 89.83 & 80.17 & 82.14 \\
SnapMix      & 70.71 & 75.76 & \underline{87.55} & \underline{90.27} & 80.77 & 83.29 \\
\midrule
Manifold     & 68.29 & 74.02 & 85.23  & 88.43 & 79.60 & 82.08 \\
MoEx         & 65.00 & 71.05 & 85.37  & 88.94 & 79.21 & 81.48 \\
RecursiveMix & 67.02 & 73.80 & 87.36  & 89.71 & 79.90 & \underline{83.35} \\
AlignMixup   & 71.78 & 76.07 & 87.18  & 89.09 & \underline{80.89}  & 82.11 \\
\midrule
\textbf{Catch-up Mix}       & \textbf{72.44} & \textbf{78.77} & \textbf{87.71} & \textbf{90.82} & \textbf{82.21} & \textbf{84.70}         \\ 
\bottomrule
\end{tabular}%
}
}
\caption{Top-1 accuracy rates (\%, $\uparrow$) on various fine-grained datasets using ResNet-18, and DenseNet-121. }
\label{tab:fine_acc}
\end{table}

\subsection{Performance on Fine-grained Datasets} 
\label{Fine-grained Visual Classification Dataset}
To show that our improvement is not limited to specific datasets, we conduct evaluations on three standard fine-grained datasets: CUB, Cars, and Aircraft.
We use ResNet-18 and DenseNet-121 as a backbone architecture. For the pre-processing, we resized input images to 256$\times$256 resolution and randomly cropped them with 224$\times$224 resolution. Next, we compare mixup methods used in Section~\ref{exp:general}. Here, we follow a hyper-parameter setting in SnapMix~\cite{huang2021snapmix}. 
As shown in Table~\ref{tab:fine_acc}, Catch-up Mix shows high performance on a variety of datasets.
With ResNet-18, Catch-up Mix achieves 72.44\%, 87.71\%, and 81.79\% on CUB, Cars, and Aircraft, outperforming the best competitor by 0.62\%, 0.10\%, and 0.90\%, respectively.

\subsection{Performance on Small Datasets} \label{data scarcity}

\begin{table}[t]
    \centering
    \resizebox{\linewidth}{!}{%
    \small{
    \begin{tabular}{lccccccc}
    \toprule
    \multirow{2}{*}{\textbf{Method}} & \multicolumn{3}{c}{\textbf{CIFAR-100}}& \multicolumn{3}{c}{\textbf{Tiny-ImageNet}}\\
    \cmidrule(lr){2-4} \cmidrule(lr){5-7}  
                 & 10\%   & 25\%  & 50\% & 10\%   & 25\%  & 50\% \\
    \midrule
    Baseline              & 35.88 & 60.17 & 70.48 & 28.09 & 43.58 & 56.13 \\
    InputMixup            & 45.65 & 63.41 & 73.19 & 30.70 & 44.85 & 55.18 \\
    CutMix                & 47.03 & 62.59 & 72.90 & 30.83 & 44.19 & 54.74 \\
    PuzzleMix             & 43.87 & 65.75 & 75.30 & 32.31 & 46.78 & 56.52 \\
    SaliencyMix           & 47.93 & 65.12 & 73.93 & 31.20 & 48.60 & 54.98 \\
    Co-Mixup              & 40.59 & 61.55 & 72.31 & 30.90 & 45.62 & 55.99 \\
    \midrule
    Manifold              & 49.00 & 64.80 & 73.46 & 33.91 & 48.53 & 56.93 \\
    MoEx                  & 41.61 & 65.03 & 72.91 & 31.58 & 48.78 & 56.46 \\
    RecursiveMix          & 36.72 & 62.94 & 72.80 & 30.84 & 45.10 & 56.21 \\
    AlignMixup            & \underline{50.15} & \underline{67.07} & \underline{75.60} & \underline{35.91} & \underline{49.78} & \underline{58.61} \\
    \midrule
    \textbf{Catch-up Mix} & \textbf{51.26} & \textbf{69.51} & \textbf{76.95} & \textbf{36.27} & \textbf{51.06} & \textbf{60.36} \\
    \bottomrule
    \end{tabular}
    }
    }
    \caption{Top-1 accuracy rate (\%, $\uparrow$) of mixup baselines trained using PreActResNet-18 with a reduced number of images per class on CIFAR-100 and Tiny-ImageNet.}\label{tab:reduced}
\end{table}

Generally, training deep neural networks for real-world applications often faces the challenge of overfitting due to insufficient data. 
Data augmentation has been considered as a key strategy to effectively increase the dataset size and alleviate this issue. 
In this section, we evaluate the performance of mixup methods under data scarcity. 
We train PreActResNet-18 on 10\%, 25\%, and 50\% subsets of the CIFAR-100 and Tiny-ImageNet, repeating each training three times with different seeds.
Training details are the same as in Table~\ref{tab:cifar_acc_corrupted} and~\ref{Tab:ti_acc_deformation_acc}.
We report the average accuracy of each method over three runs in Table~\ref{tab:reduced}. 
The results show that Catch-up Mix effectively prevents the overfitting phenomenon and achieves greater performance even when there's a limited number of data per class.
Notably, RecursiveMix exhibits inferior performance compared to other mixup methods. 
We hypothesize that this is due to its reliance on a contrastive learning framework, which requires high data diversity to learn meaningful representations.

\subsection{Performance on Large Dataset}

\begin{table}
    \centering
    % \resizebox{0.98\columnwidth}{!}{
    \small{
        \begin{tabular}{lcccccc}
            \toprule
            \multirow{3}*{\textbf{Method}}  & \textbf{ImageNet} & \multicolumn{3}{c}{\textbf{ImageNet-O}} \\
            \cmidrule(lr){2-2} \cmidrule(lr){3-5} 
            & Acc & FPR95 & AUROC & AUPR \\
            & (\%, $\uparrow$) & (\%, $\downarrow$) & ($\uparrow$) & ($\uparrow$) \\
            \midrule
            Baseline   & 76.15 & -*    & -*    & -*          \\
            Mixup      & 77.46 & 90.06 & 53.55 & 16.27       \\
            CutMix     & 78.58 & 93.53 & 51.19 & 15.81       \\
            PuzzleMix  & 78.63 & 91.08 & 53.63 & 16.58       \\
            ManifoldMix   & 77.50 & 91.66 & 52.99 & 16.23       \\
            MoEx & \underline{79.06} & 93.99 & 51.58 & 15.99 \\
            RecursiveMix & \textbf{79.14} & \underline{88.54} & \underline{54.99} & \textbf{16.91} \\
            \midrule
            \textbf{Catch-up Mix} & 78.71 & \textbf{81.92} & \textbf{55.56} & \underline{16.69} \\
            \bottomrule
        \end{tabular}
    }
    \caption{
    Top-1 classification accuracy rate (\%) on ImageNet-1k with ResNet-50 and out-of-distribution detection performance (FPR95, AUROC, and AUPR) on ImageNet-O.
    *ImageNet-O consists of data that vanilla ResNet misclassifies, meaning its detection accuracy is 0\%. 
    Here, to perform OOD evaluation, Catch-up Mix is compared with methods that publicly released their pre-trained model weights.
    } 
    \label{tab:imagenet}
    % }
\end{table} 

To evaluate our method in a data-rich context, we train ResNet-50 on ImageNet-1k for 300 epochs using Catch-up Mix and compare it with mixup methods that published the ImageNet pre-trained model weights. Catch-up Mix achieves a top-1 accuracy of 78.71.
Our motivation aims to leverage the under-exploited model capacity, often left unused due to the early convergence.
For expansive datasets such as ImageNet-1K, the inherent diversity and variety within the data naturally promote the model's generalization capabilities, diminishing the potentially unused capacity and the chances of early convergence.

\paragraph{Robustness against OOD} We also verify that our feature diversity and robustness improvements are valid for the ImageNet-trained model using ImageNet-O, out-of distribution (OOD) detection dataset~\cite{hendrycks2021natural}.
OOD detection tasks can provide insights into the generalization capability and feature representations of models. 
From ImageNet-22K examples with the ImageNet-1K class removed, ImageNet-O consists of images that vanilla ResNet-50 confidently classifies as belonging to one of the ImageNet-1K classes.
OOD detection performance in Table~\ref{tab:imagenet} shows that our method can effectively differentiate between the data it was trained on (in-distribution) and unfamiliar data (OOD).
Having the advantage of a variety of features, Catch-up Mix improves the robustness to unseen data and builds a reliable model.

\paragraph{Accuracy-Robustness Trade-off} 
Even when there is plenty of data and little surplus capacity in the model, we encourage a model to utilize diverse features and, if necessary, sacrifice accuracy for robustness (Table~\ref{tab:imagenet}).
When training ResNet-50 on simpler classification tasks like CIFAR-100 and considering the model's capacity (e.g., number of parameters), encouraging feature diversity can enhance both robustness and accuracy without sacrificing performance.
Conversely, when training on larger datasets for more challenging tasks such as ImageNet, prioritizing feature diversity might result in a trade-off between accuracy and robustness.

\begin{table}[t]
\centering
        % \resizebox{0.98\columnwidth}{!}{%
        \small{
        \begin{tabular}{lccccc}
            \toprule
             & \multicolumn{3}{c|}{\textbf{CIFAR-100}} & \multicolumn{2}{c}{\textbf{TI}}\\%\multicolumn{2}{c}{\textbf{Tiny-ImageNet}}\\
            \cmidrule(lr){2-4} \cmidrule(lr){5-6} 
            Model & R18* & WRN16 & \multicolumn{1}{c|}{RX50*} & R18* & PR50 \\
            (Epoch) & (400) & (400) & \multicolumn{1}{c|}{(400)} & (400) & (1200) \\
            \midrule
            Baseline     & 77.73 & 79.37$\dagger$ & 80.24 & 61.68 & 62.32 \\
            MixUp        & 79.34 & 80.12$\dagger$ & 82.54 & 63.86 & 66.51 \\
            CutMix       & 79.58 & 80.29$\dagger$ & 78.52 & 65.53 & 67.02 \\
            PuzzleMix    & 80.82 & 80.75$\dagger$ & 82.84 & 65.81 & \underline{71.39} \\
            SaliencyMix  & 79.64 & 80.41$\dagger$ & 78.63 & 64.60 & 66.37 \\
            Co-Mixup     & \underline{80.87} & 80.43$\dagger$ & \underline{82.88} & 65.92 & 65.26 \\
            \midrule
            Manifold     & 80.18 & 80.77$\dagger$ & 82.56 & 64.15 & 66.91 \\
            MoEx         & N/A   & 79.31 & N/A   & N/A   & 68.98 \\
            RecursiveMix & N/A   & \underline{81.27} & N/A   & N/A   & 68.45 \\
            AlignMixup   & 80.80 & 81.23$\dagger$ & N/A   & \underline{66.87} & 69.36 \\
            \midrule
            \textbf{Catch-up Mix} & \textbf{82.10} & \textbf{81.64} & \textbf{83.56} & \textbf{68.84} & \textbf{72.58} \\
            \bottomrule
        \end{tabular}
        % }
        }
        \caption{Top-1 accuracy rates (\%) on CIFAR-100 and Tiny-ImageNet using various architectures. 
        All the results of * refer to \cite{li2022openmixup}. 
        The results of WRN16$\dagger$ are from~\cite{venkataramanan2022alignmixup}.
        }
\label{tab:various_arch}
\end{table}

\subsection{Performance on Various Architectures} \label{Various Architecture}
Here, we evaluate Catch-up Mix on various architectures to show that our performance gain is not limited to specific architectures. 
We evaluate ResNet-18 (R18), Wide ResNet-16-8 (WRN16), and ResNeXt-50 (RX50) on CIFAR-100 and evaluate ResNet-18 (R18) and PreActResNet-50 (PR50) on Tiny-ImageNet.
Experiment settings for R18 and RX50 were adopted from \citet{li2022openmixup}, while for WRN16 and PR-50, we referenced Co-Mixup~\cite{kim2020co} and AlignMixup~\cite{venkataramanan2022alignmixup}.
E.g., we train PR-50 for 1200 epochs and the others for 400 epochs.
As depicted in Table~\ref{tab:various_arch}, Catch-up Mix consistently outperformed other mixup methods in various architectures.

\subsection{Loss Landscape}

\begin{figure}[t]
\centering
\includegraphics[width=0.97\columnwidth]{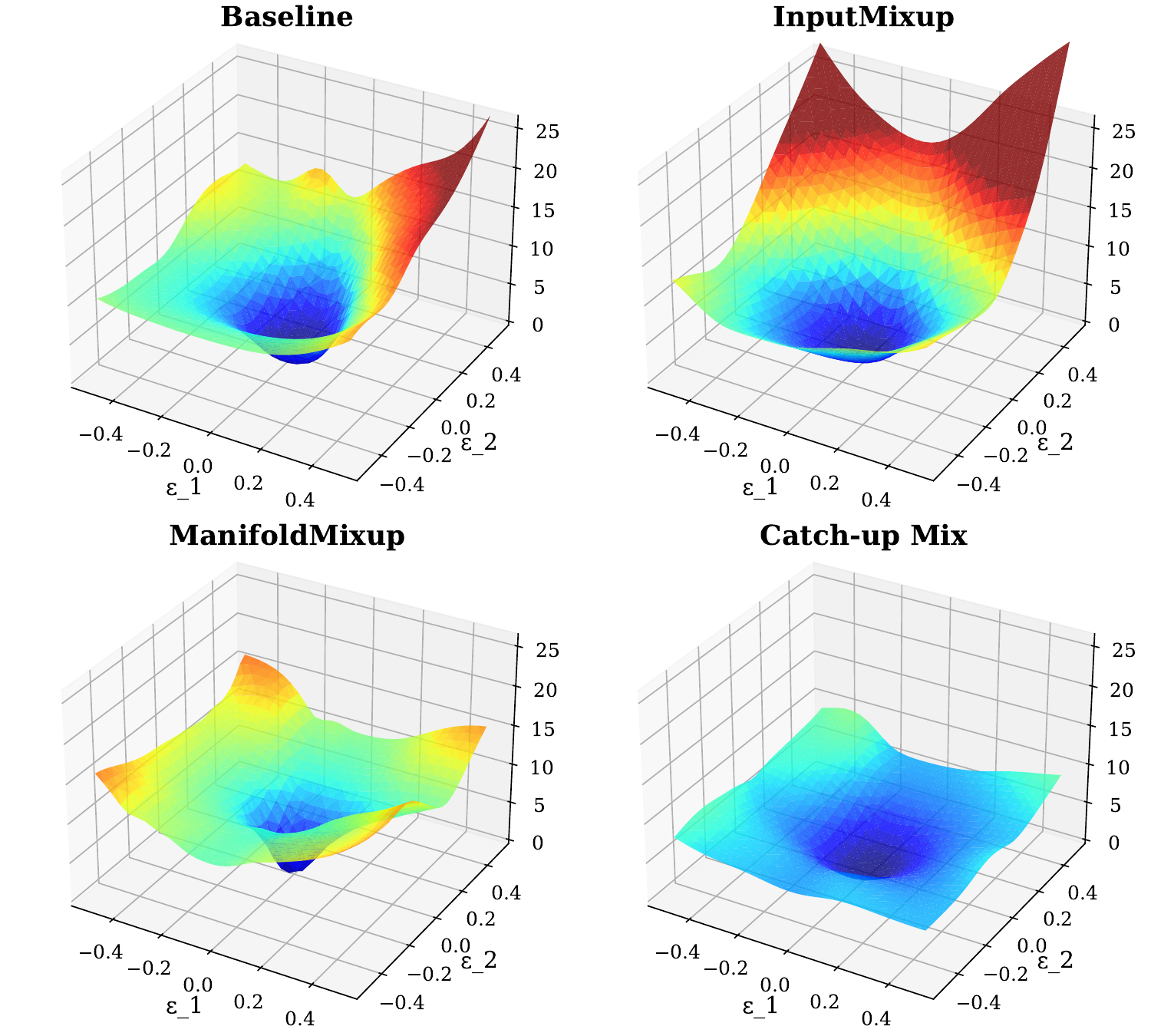}
\caption{Loss Landscapes of mixup methods on Tiny-ImageNet using PreActResNet-18. }\label{fig:landscape}
\end{figure}

The concept of generalization in a neural network pertains to the model's ability to effectively apply the knowledge obtained from the training data to unseen test data. 
A well-generalizing model reduces the discrepancy called the `generalization gap' between the performance during training and testing. 
In this context, numerous studies~\cite{keskarlarge,foretsharpness,choromanska2015loss} have discussed that a network's convergence to flat minima enhances its generalization and robustness performance. 
To this end, the method generates a flatter loss landscape compared to others would be better generalization and robustness. 
Following the implementation of~\citet{yao2020pyhessian}, we plot the loss landscape of mixup methods as shown in Figure~\ref{fig:landscape}. 
We use InputMixup~\cite{zhang2018mixup} as the input-level mixup except for `Baseline' for the fair comparison. 
Loss values of other mixup methods change extremely, even with a slight change. However, interestingly, the model trained with Catch-up Mix has a flat loss landscape, and its loss value does not explode with small changes compared to InputMixup and ManifoldMixup. 
As intuitively shown by the loss landscape, Catch-up Mix consistently achieves high robustness and generalization on various benchmarks.

\section{Conclusion and Future Work}
We have addressed the challenge of models relying heavily on a small subset of filters for making predictions, resulting in limited generalization and robustness. 
We introduced Catch-up Mix, a technique that enhances learning opportunities for filters that may fall behind during training. 
By mixing activation maps with lower norms, Catch-up Mix encourages the development of more diverse representations and mitigates over-reliance problems. 
Our experiments highlight Catch-up Mix's improved performance on vision classification datasets, enhancing robustness against adversarial attacks, data corruption, and deformations.

Previous studies have not investigated the model's unused capacity during training. 
Our approach aims to bridge this gap, moving from a state where a significant portion of the model's capacity remains idle to an optimized state where every filter actively contributes to predictions.
Therefore, this enhancement is more pronounced when the dataset and regularizations cannot sufficiently utilize the model's capacity. 
However, a challenge arises when all the weights can already be fully optimized (e.g., plenty of training data); utilizing various features becomes a cost for the generalization, resulting in an accuracy-robustness trade-off.
In future work, our research will address these considerations, further improving the performance and adaptability of Catch-up Mix.

\section*{Acknowledgements}
This work was supported by Institute of Information \& communications Technology Planning \& Evaluation (IITP) grant funded by the Korea government (MSIT) (No.2021-0-00456, Development of Ultra-high Speech Quality Technology for remote Multi-speaker Conference System), and by the Korea Institute of Science and Technology (KIST) Institutional Program.

\bibliography{aaai24}

\clearpage
\appendix
\section{Appendix}
\subsection{Catch-up Mix Process in Algorithm}
Throughout the paper, for ease of understanding, we describe the overall Catch-up Mix process as if it were done for both inputs.
However, in implementation, it is performed in batch incurring minimal overhead, and does not require additional forward and backward propagation.
To illustrate this as a batch-wise operation in Algorithm~\ref{alg:Catch-up Mix}, we first obatain feature map $h \in \mathbb{R}^{B \times C \times H \times W}$  from the selected layer, and use it as source feature.
Then, target feature map $h'$ is given as shuffling the index of $h$ in batch dimension.
\begin{equation}
h' = h[\text{shuffled\_index}, :, :, :]
\end{equation}
This means that $h$ and $h'$ consist of the same feature map with different order.

% For better understanding, we describe the overall Catch-up Mix process in Algorithm~\ref{alg:Catch-up Mix} as its process for each pair of inputs. 
% % Note that $h$ and $h'$ are different feature maps. 
% Extending this to a batch-wise operation, $h'$ is selected from the mini-batch by shuffling the index of $h$. This means that $h$ and $h'$ consist of the same feature map with different order.
% In the implementation, Catch-up Mix is performed in batches incurring minimal overhead. 
% % After drawing the mixed feature, it forwards the rest of layers and compute classification loss for update.

\begin{algorithm}[ht]
%\setstretch{1.4} %require \usepackage{setspace}
    \caption{Catch-up Mix procedure}\label{alg:Catch-up Mix}
    \begin{algorithmic}[1]
    %\in \mathbb{R}^{H \times W \times C}$, \in \mathbb{R}^{N}$
        \Require{Feature map $h, h'$ , Label $y, y'$, mixing ratio $\lambda \sim Beta(\alpha,\alpha)$} 
        \Ensure{Mixed feature $h_{mix}$, Mixed label $y_{mix}$}
        \vspace{6pt}
        \Function{Catch-up Mix}{$h,h',y,y',\lambda$}
            \State $FI\; \gets \{FI_{c}\;|\;FI_{c}\;=\;||h_{c}||_{2}\quad :\forall c \in C\}$ 
            \State $FI' \gets \{FI'_{c}\;|\;FI'_{c}\;=\;||h'_{c}||_{2}\quad :\forall c \in C\}$  
            \Statex \Comment{Compute feature influence $FI,FI' \in \mathbb{R}^{C}$}
            \vspace{3pt}
            \State $RFI \gets \; \frac{FI}{\sum_{c=0}^{C}(FI_c)} \;-\; \frac{FI'}{\sum_{c=0}^{C}(FI_c')}$  
            \vspace{3pt}
            \Statex  \Comment{Compute Relative $FI$ ($RFI$)}
            \State $N_{mix} \gets \lfloor{\lambda \times |C|} \rfloor$ 
            \State $\mathcal{C}_{retain} \gets \{c\;|\;c\;\in\; \underset{c}{ArgTopK}(-RFI,N_{mix})\}$ 
            \vspace{3pt}
            \Statex   \Comment{Choose $N_{mix}$ channels with low RFI}
            \vspace{6pt}
            \State $\mathbb{M}\gets\{m_{c}|m_{c}=
                \begin{Bmatrix}  
                    \mathbf{1}_{hw},  & if\; c \in \mathcal{C}_{retain} \\ 
                    \mathbf{0}_{hw},  & otherwise \\
                \end{Bmatrix}\}$ 
            \vspace{6pt}
            \Statex   \Comment{Mixing mask generation}
            \vspace{3pt}
            \State $h_{mix} \gets \mathbb{M} \odot h + (\mathbb{I} - \mathbb{M}) \odot h'$  \Comment{Feature mixing equation}
            \State $y_{mix} \gets \lambda \times y + (1-\lambda) \times y'$ \Comment{Label mixing equation}
        \State \Return $h_{mix},\;y_{mix}$ 
        \EndFunction
    \end{algorithmic}
\end{algorithm}

\begin{figure}[ht]
    \centering
    \begin{subfigure}[b]{0.49\columnwidth}
        \includegraphics[width=\linewidth]{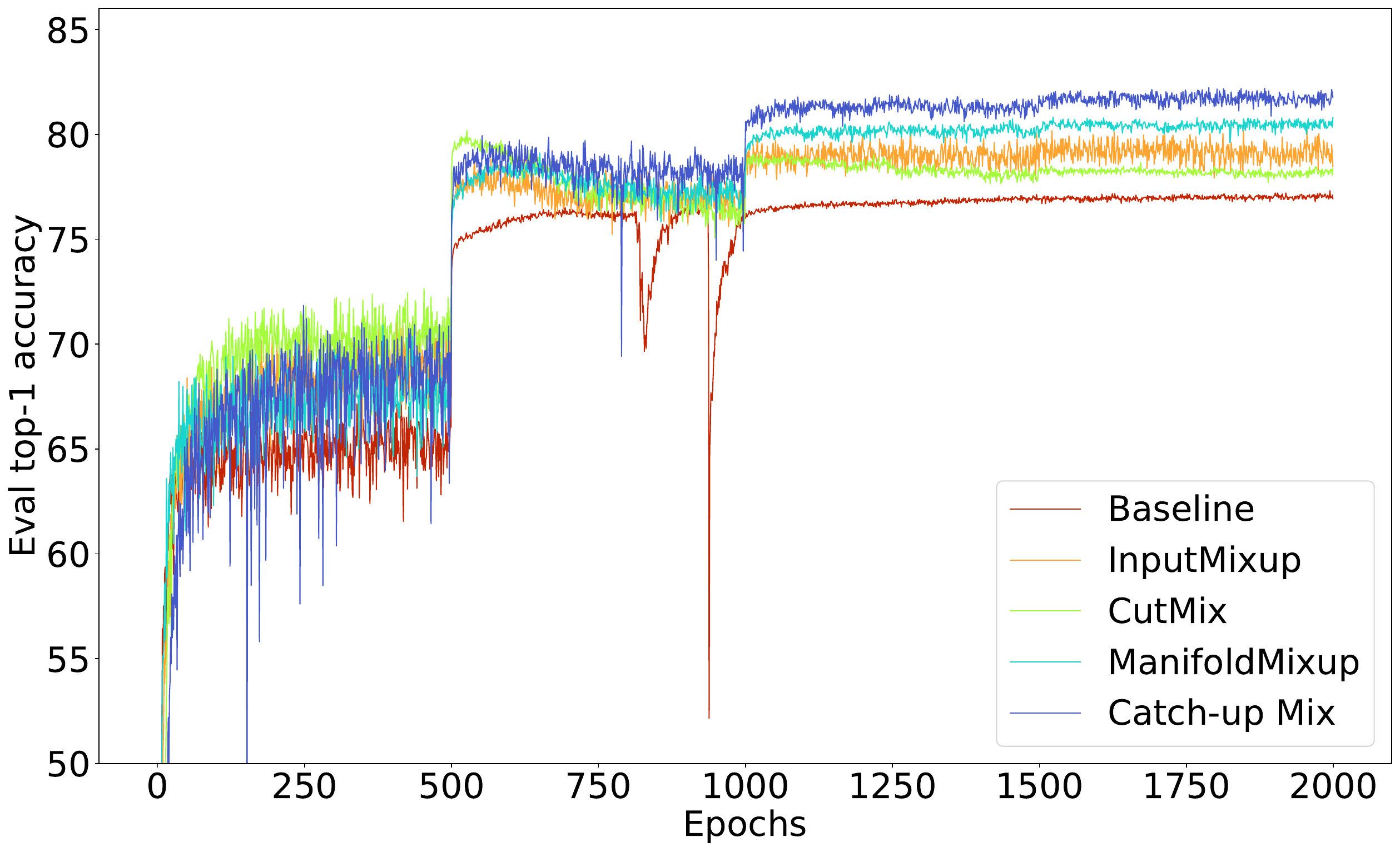}
        \vspace{-0.5cm}
        \caption{Validation top-1 accuracy (CIFAR)}
        \label{fig:appendix_cifar_acc}
    \end{subfigure}
    \hfill
    \begin{subfigure}[b]{0.49\columnwidth}
        \includegraphics[width=\linewidth]{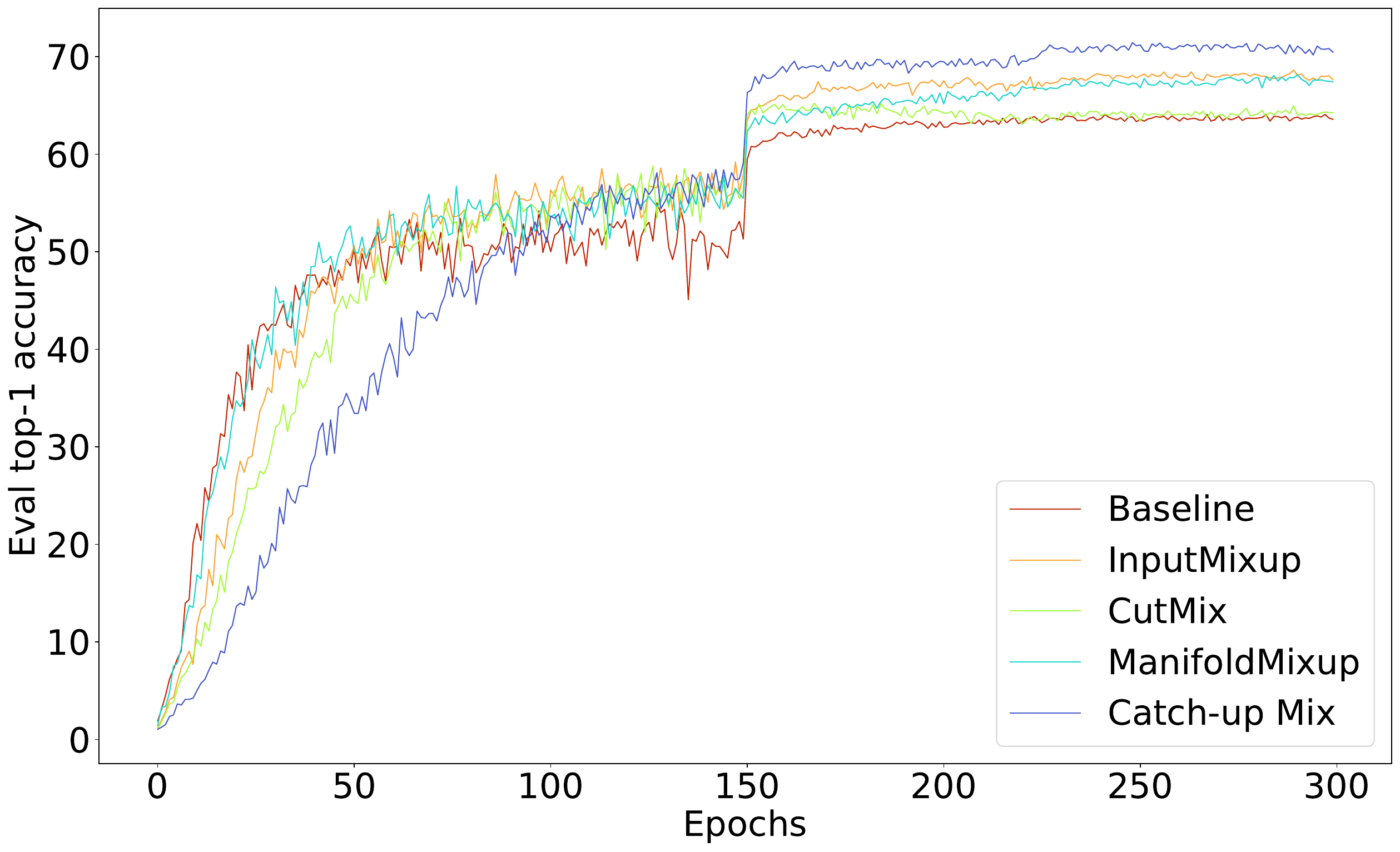}
        \vspace{-0.5cm}
        \caption{Validation top-1 accuracy (CUB)}
        \label{fig:appendix_cub_acc}
    \end{subfigure}
    \vskip\baselineskip
    \begin{subfigure}[b]{0.49\columnwidth}
        \vspace{-0.1cm}
        \includegraphics[width=\linewidth]{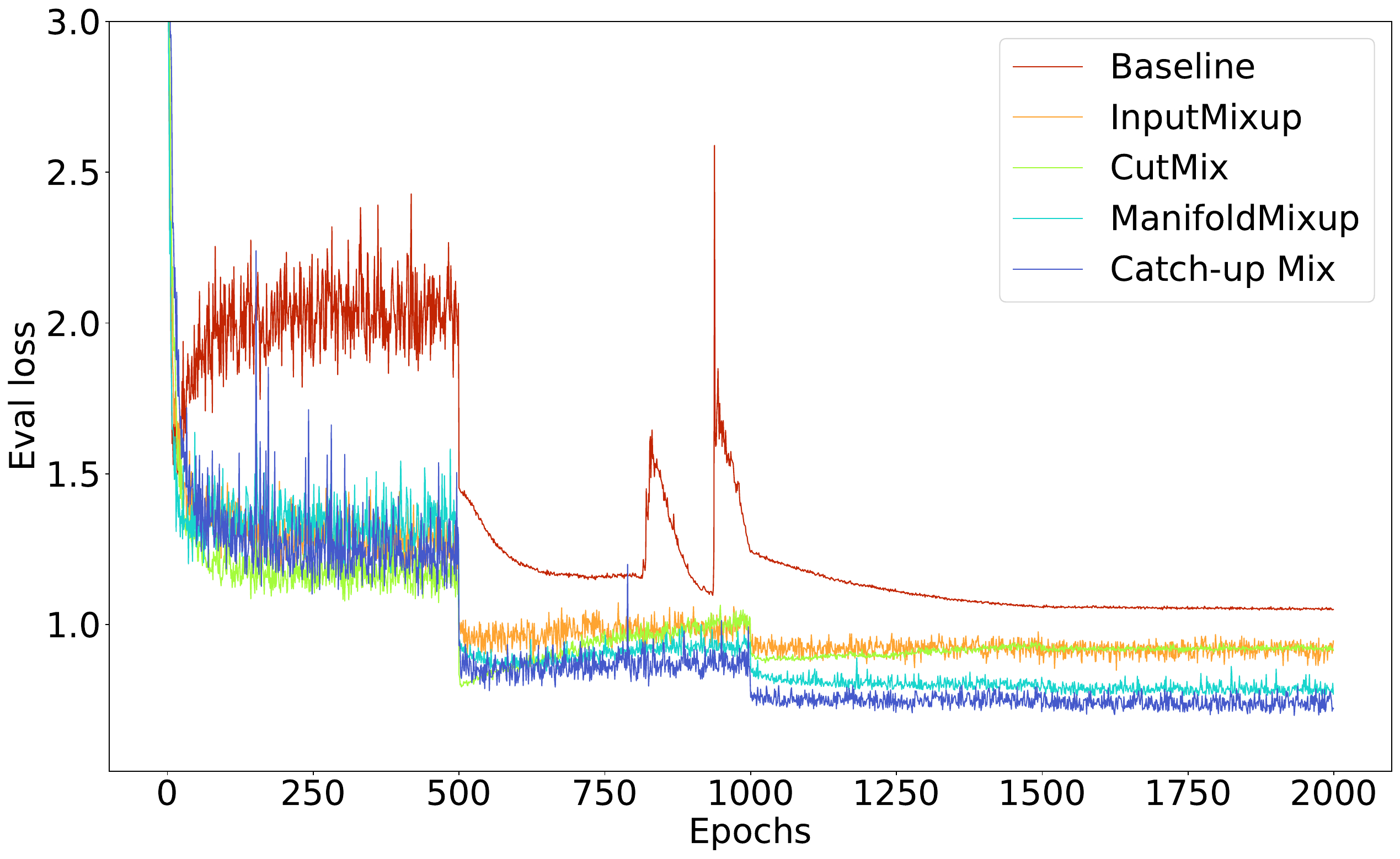}
        \vspace{-0.5cm}
        \caption{Validation loss (CIFAR)}
        \label{fig:appendix_cifar_loss}
    \end{subfigure}
    \hfill
    \begin{subfigure}[b]{0.49\columnwidth}
        \vspace{-0.1cm}
        \includegraphics[width=\linewidth]{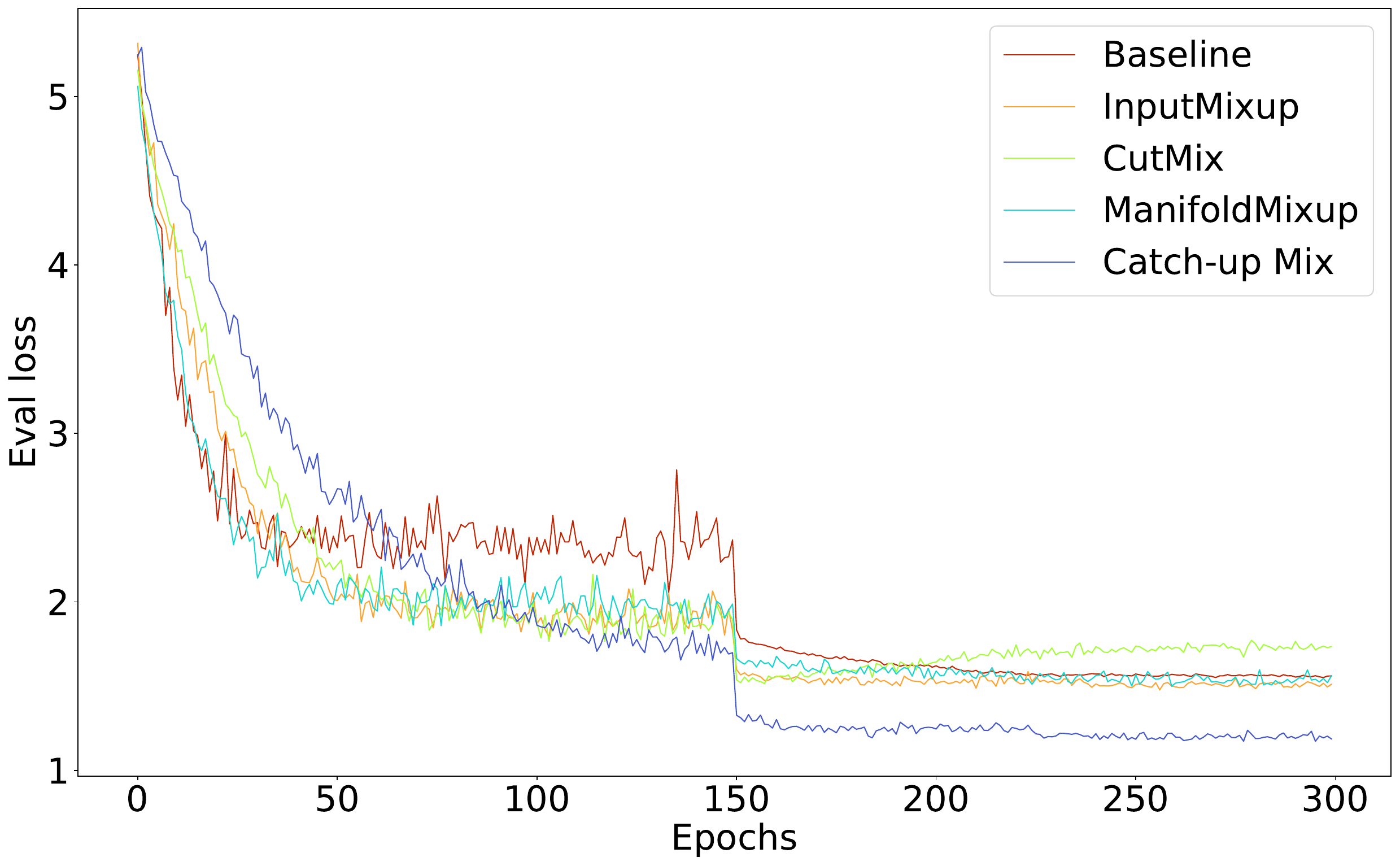}
        \vspace{-0.5cm}
        \caption{Validation loss (CUB)}2
        \label{fig:appendix_cub_loss}
    \end{subfigure}
    \vskip\baselineskip
    \begin{subfigure}[b]{0.49\columnwidth}
        \vspace{-0.1cm}
        \includegraphics[width=\linewidth]{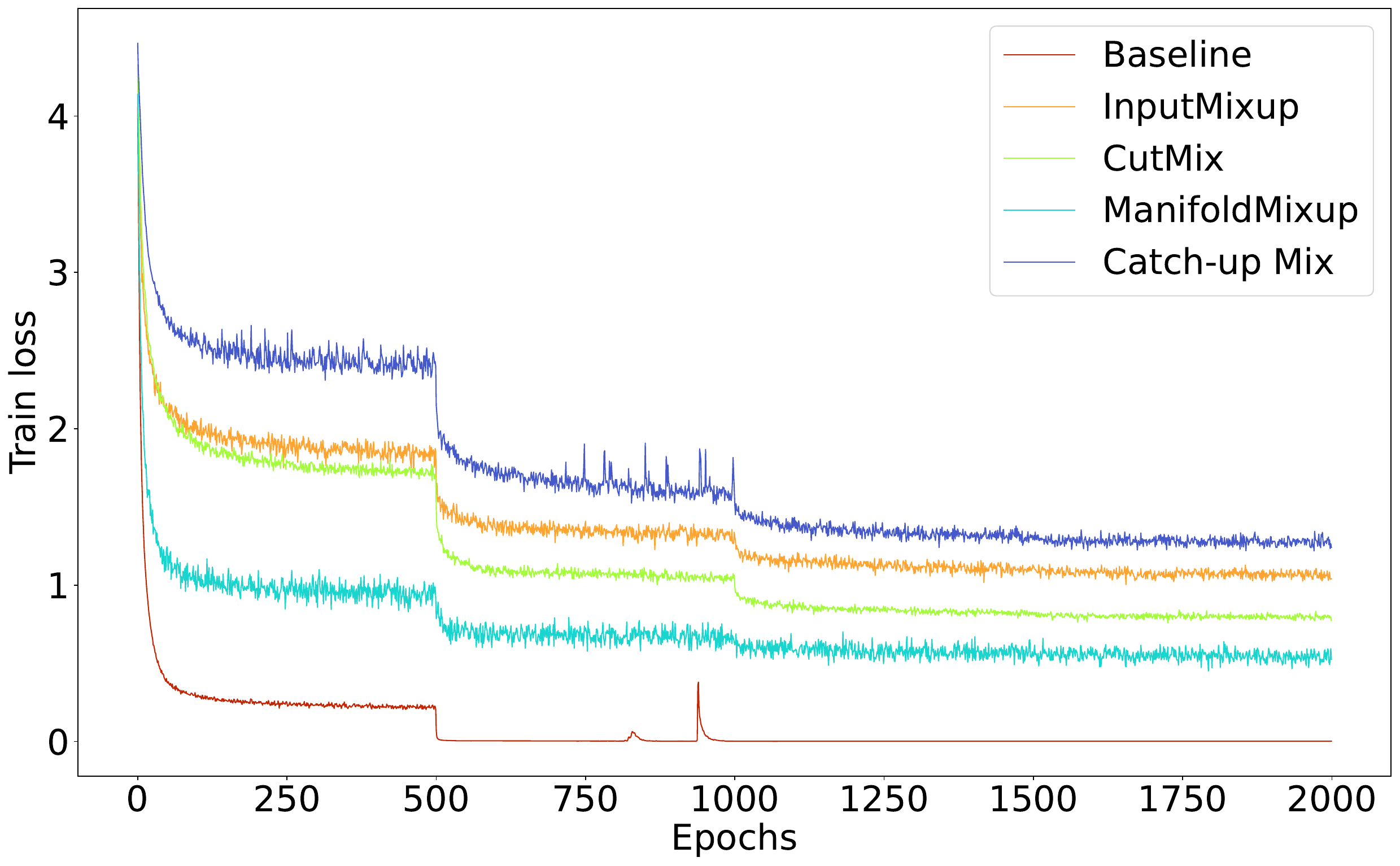}
        \vspace{-0.5cm}
        \caption{Training loss (CIFAR)}
        \label{fig:appendix_cifar_train_loss}
    \end{subfigure}
    \hfill
    \begin{subfigure}[b]{0.49\columnwidth}
        \vspace{-0.1cm}
        \includegraphics[width=\linewidth]{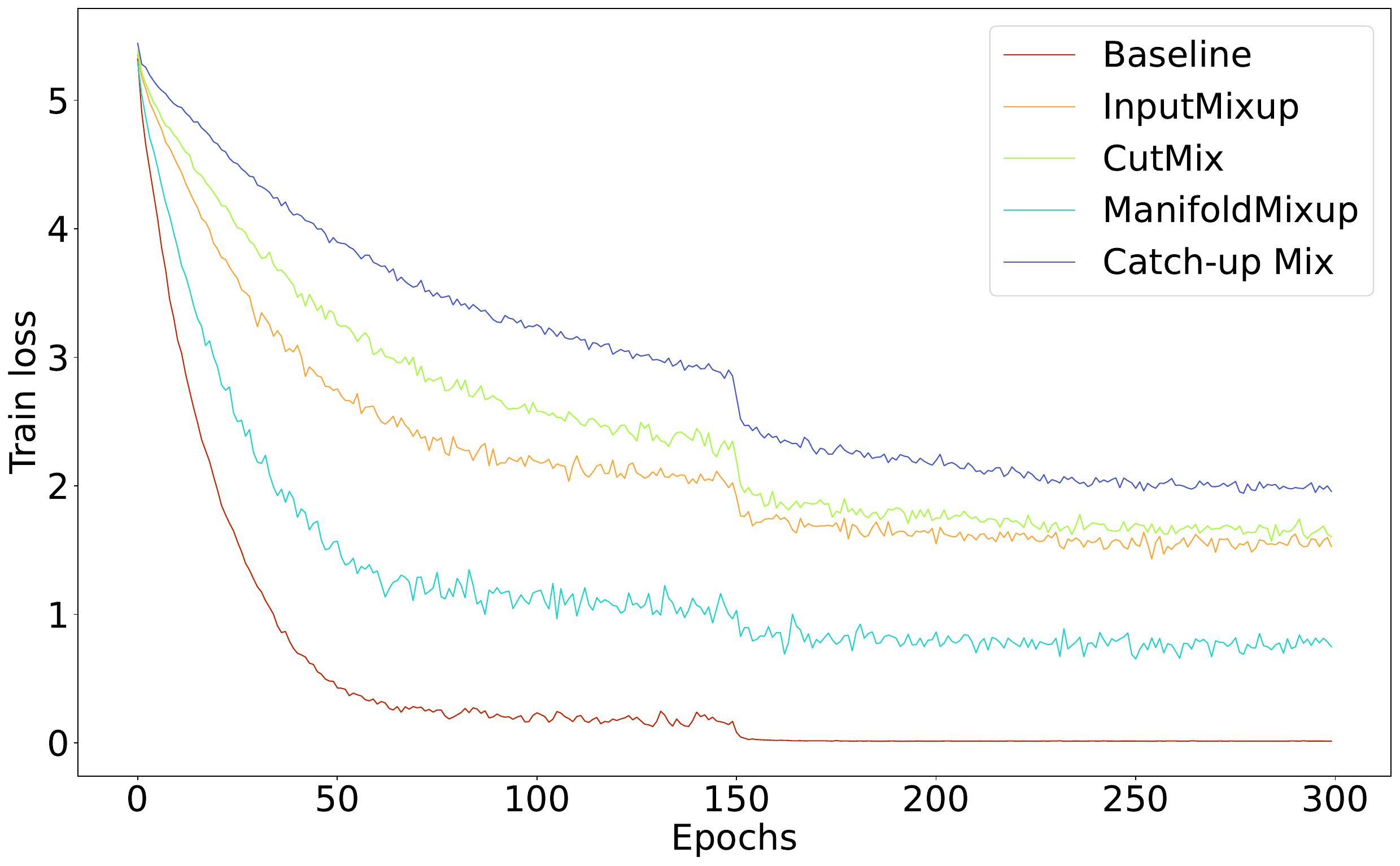}
        \vspace{-0.5cm}
        \caption{Training loss (CUB)}
        \label{fig:appendix_cub_train_loss}
    \end{subfigure}
    \caption{Top-1 accuracy rates (\%), validation loss, and training loss curve of mixup baselines on CIFAR using PreActResNet-18 and CUB using ResNet-18. } \label{fig:appendix_acc_loss}
\end{figure}

\begin{figure*}[ht]
    \centering
    \includegraphics[width=0.99\linewidth]{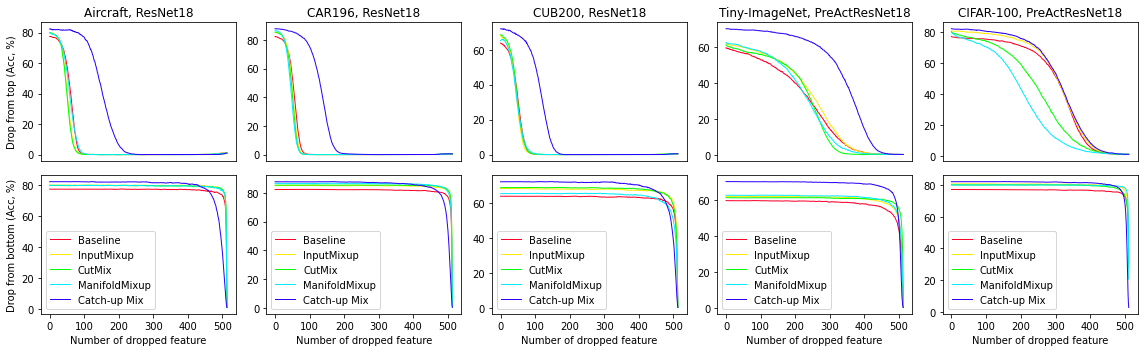}
    \caption{Correlation Between feature reduction and model accuracy. After removing each dimension of the latent vector with the highest and the lowest values, we measure the top-1 accuracy by making predictions with the remaining features.}
    \label{fig:FeatureReliance}
\end{figure*}

\subsubsection{Analysis on Loss and Accuracy Curve} 
In this section, we analyze the loss and accuracy curves in the process of training ResNet with various mixup methods (Figure \ref{fig:appendix_acc_loss}).
We compare the validation loss on CUB~\cite{WahCUB_200_2011} using ResNet-18 and train the model for 300 epochs following the experimental settings described in \citet{huang2021snapmix}, and CIFAR-100 (CIFAR)~\cite{Krizhevsky2009learning} using PreActResNet-18 for 2000 epochs following the protocol of \citet{venkataramanan2022alignmixup}. 

The model with an overfitting problem typically results in low training loss but high validation loss due to poor generalization of unseen data.
Our method, Catch-Up Mix, deviates from such overfitting scenarios. 
This is accomplished by utilizing a mixed feature map derived from the activations of less mature filters.
Consequently, the model is presented with more challenging samples during training, contributing to the increased training loss. 
However, since the model learns by utilizing a variety of filters rather than relying on a small number of filters, the validation loss is lower, reflecting an improved ability to generalize to unseen data. 
Examining the validation loss and accuracy curves, Catch-Up Mix appears to learn features steadily at a relatively slow pace compared to the other methods, especially for ResNet-18. 
We believe that this slower but deeper convergence is a result of Catch-Up Mix promoting a more balanced development of convolutional filters.

\subsection{Over-Reliance and Feature Diversity} 

This section explores whether the trained model depends on only a few features or employs a diverse range of features to classify objects.
To do this, we extract latent vectors from the evaluation dataset, specifically from the output layer immediately preceding the final fully connected layer. 
These latent vectors are lower-dimensional representations of input images, known as the feature space. 
For instance, in ResNet-18~\cite{he2016deep} and PreActResNet-18~\cite{he2016identity}, each image corresponds to a 512-dimensional latent vector. 
These dimensions encapsulate underlying data aspects or structures, such as edges, textures, shapes, or object parts essential for distinguishing between classes. 
Thus, in this section, we refer to each dimension of the latent vector as a `\textit{feature}.'

Subsequently, we assess the model's over-reliance problem and feature diversity by monitoring the accuracy decline as we progressively eliminate features with higher magnitude (see Figure \ref{fig:FeatureReliance}). 
A steep accuracy drop indicates an over-reliance on a limited set of features.
In our experiments, Catch-Up Mix demonstrates robust performance even upon removal of key features, suggesting that it extracts and utilizes a broader spectrum of features from the input data. 
Conversely, dropping features from the lowest value had little impact on accuracy, even when more than 80\% of the latent vector values disappeared.
This observation indicates that features with high values have a high impact on prediction, and it also shows that only a small number of features are utilized for model prediction.

\begin{figure}[ht]
    \centering
    \includegraphics[width=0.6\linewidth]{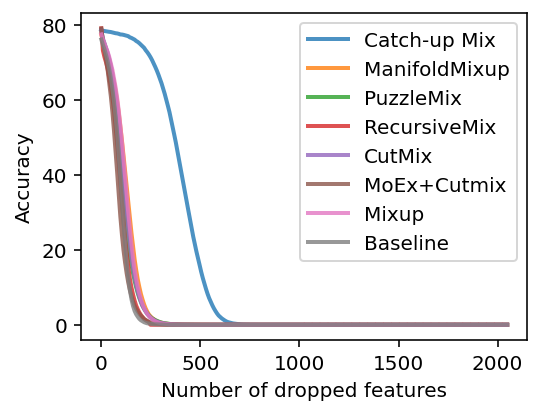}
    \caption{Feature Reliance (drop from top) for ImageNet-1k dataset on ResNet-50.}
    \label{fig:FeatureReliance_imagenet}
\end{figure}

We evaluate the model's tendency to over-rely on certain features and its feature diversity by observing the accuracy trajectory as we gradually remove  features of the highest magnitude (refer to Figure \ref{fig:FeatureReliance}).
Additionally, we found similar dependency issues with imageNet trained models in Figure \ref{fig:FeatureReliance_imagenet}.
A sharp accuracy drop indicates an excessive reliance on a small number of features .
In our experiments, Catch-Up Mix consistently maintains its performance, even in the absence of salient features. 
This suggest that it extracts and utilizes a broader spectrum of features from the input data.
Conversely, dropping the least significant features had a negligible effect on accuracy, even with a reduction exceeding 80\% of the latent vector.
This finding reveals that only a few features are used for model prediction and that features with high values have a significant impact on prediction.

\begin{table}[ht]
    \small
    \centering
    \begin{tabular}{lccc}
    \toprule
     \multirow{2}*{\textbf{Method}} & GT-known & Top-1    & Top-1  \\
                                    & Loc (\%) & Loc (\%) & Acc (\%) \\
    \midrule
    Baseline      & 62.71  & 48.66  & 76.66 \\
    InputMixup    & 54.15  & 39.59  & 76.84 \\
    CutMix        & \underline{64.08} & \underline{49.93}  & 74.70 \\
    SnapMix       & 60.12  & 46.46  & \textbf{78.43} \\
    \midrule
    \textbf{Catch-up Mix} & \textbf{66.72}  & \textbf{51.51}  & \underline{76.94} \\
    \bottomrule
    \end{tabular}%
    \caption{Weakly supervised object localization results. GT-known localization, top-1 localization, and top-1 accuracy rate (\%, $\uparrow$) on CUB using pre-trained ResNet architecture.} \label{tab:loc}
\end{table}

\subsection{Weakly Supervised Object Localization}

Weakly Supervised Object Localization (WSOL) techniques aim to learn the location of an object without any location annotation but with classification labels. Generally, existing approaches utilize a pre-trained model to fine-tune the target dataset to evaluate localization performance. 
Selective region discarding techniques have been proposed in previous studies \cite{choe2019attention, singh2017hide} to improve object localization. However, these approaches often face a trade-off between classification accuracy and localization performance during the fine-tuning process.
Inspired by CutMix~\cite{yun2019cutmix}, which reinforces identifying the object from a partial view effectively by replacing the patch, Catch-up Mix can also fully utilize the model's capability because it makes predictions and updates by mixing less developed filters for the input. It improves the issue of the model being overly dependent on specific filters and enhances the extraction capability of the filters which require catch-up class. To validate the localization performance, we use pre-trained ResNet-18 to fine-tune on 224$\times$224 resized CUB-200-2011~\cite{WahCUB_200_2011}. We first compute an image's class activation map (CAM) \cite{zhou2016cam} and binarize it with a specific threshold of 0.2 for all our experiments to discriminate foreground and background regions in CAM. Then, with a bounding box from the binary map, we measure the top-1 (Top-1 Loc) and ground-truth known (GT-known Loc) localization accuracy rate (\%, $\uparrow$). GT-known Loc confirms an answer as correct when the intersection over union (IoU), which is a measure of overlap between an estimated bounding box by CAM and a ground truth bounding box for the actual class, equals or exceeds 50\%. Top 1 Loc is the metric that rises when correct on both classification and GT known localization. As shown in Table \ref{tab:loc}, Catch-up Mix outperforms both other mixup methods without observing a trade-off between localization and classification.

\section{Ablation Study}

\subsection{Catch-up Mix with Various Input-level Mixup Methods} 
\label{appendix_sec:inputlevel}
In the main paper, we present the results of utilizing Catch-up Mix by randomly selecting layers and applying CutMix~\cite{yun2019cutmix} when the input layer is selected.
Specifically, we sample $k$ from the set $K$=\{0, 1, 2, 3, 4, 5\}, resulting in a probability of 1/6 for applying CutMix. 
To substantiate that the model performance is not solely attributable to CutMix, we also report the performance utilizing a different mixup strategy (InputMixup~\cite{zhang2018mixup}), as well as without any input-level mixup. 
This comprehensive evaluation allows us to assess the impact of varying mixup techniques on the overall model performance. 
All the input-level mixup we evaluate have hyper-parameter $\alpha$ for determining mixing ratio $\lambda \sim Beta(\alpha,\alpha)$. In addition, we use an official hyper-parameter, which is $\alpha$ for CutMix of 1 and InputMixup of 0.2.
Despite slight differences attributed to the choice of input-level mixup method, our approach consistently exhibits significant enhancements in performance (see Table \ref{ablation:input}).

\begin{table}[t]
\centering
    % \resizebox{\columnwidth}{!}{%
    \small
    \begin{tabular}{lcc}
        \toprule
        Method          & CIFAR-100     & Tiny-ImageNet \\
        \midrule
        No Mixup        & 81.48 (0.46)  & 68.44 (0.32) \\
        with InputMixup & 81.84 (0.50)  & 68.58 (0.18) \\
        with CutMix     & 82.24 (0.04)  & 69.27 (0.09) \\
        \bottomrule
    \end{tabular}%
    \caption{
    Top-1 accuracy (\%, $\uparrow$) of Catch-up Mix with different input-level mixup on CIFAR-100 and Tiny-ImageNet using PreActResNet-18.
    We experiment with three different seeds and report average and standard deviation of accuracy.}
    \label{ablation:input}
    % }
\end{table}  

\begin{figure}[t]
    \includegraphics[width=\columnwidth]{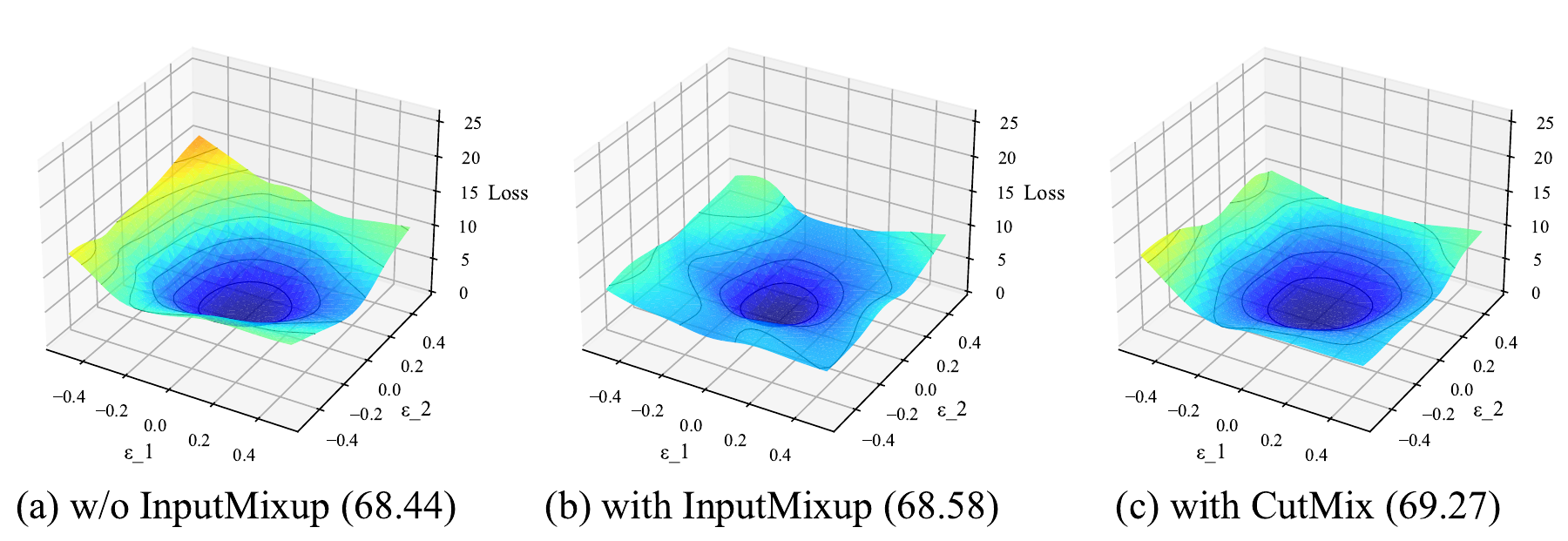}
    \caption{
    Loss Landscapes of Catch-up Mix using different input-level mixup methods.
    }
    \label{fig:appendix_landscape}
\end{figure}

\paragraph{Loss Landscape with Various Input-level Mixup Methods} 
We also visualize the loss landscape of the model trained using Catch-up Mix (a) without input-level mixup, (b) with InputMixup~\cite{zhang2018mixup}, and (c) with CutMix~\cite{yun2019cutmix}). 
We use PreActResNet-18 trained on Tiny-ImageNet.
Since input-level mixup methods slightly affect the model performance and robustness, these also affect the form of loss landscape. 
To this end, we compare the loss landscape of the model trained with Catch-up Mix using different input-level mixup methods in order to validate that our method itself contributes to flattening the loss landscape. 
As illustrated in Figure \ref{fig:appendix_landscape}, our Catch-up Mix guides the model to obtain a flattened loss landscape and does not highly depend on input-level mixup methods.

\subsection{Feature Selection Strategy}

Catch-up Mix operates by comparing the activation norms of each filter on randomly paired data and generating a mixed feature map consisting of activations with smaller norms. 
To validate the effectiveness of our mixup feature selection strategy, we compare it to custom baseline for channel mixup, which randomly mixes the activation map, which we denote as `RandomChannelMix.'
We trained PreActResNet-18 with CIFAR-100 for 2000 epochs and Tiny-ImageNet for 1200 epochs same as in Tables 1,2 of main paper.
Here, we do not perform input-level mixup for RandomChannelMix and Catch-up Mix.
From Table \ref{ablation:randomchannelmix}, we find that channel selection of activation maps without any strategy does not help to regularize the model but rather harms the generalization performance of the model. 

\begin{table}[t]
\centering
\small
    \begin{tabular}{lcc}
        \toprule
        Method           & CIFAR-100 & Tiny-ImageNet\\
        \midrule
        Baseline         & 77.02     & \underline{63.02} \\
        RandomChannelMix & \underline{77.30}     & 61.21 \\
        Catch-up Mix     & \textbf{81.48}     & \textbf{68.44} \\
        \bottomrule
    \end{tabular}%
    \caption{Top-1 accuracy (\%, $\uparrow$) of Baseline, RandomChannelMix, and Catch-up Mix using PreActResNet-18 on CIFAR-100 and Tiny-ImageNet. For RandomChannelMix and Catch-up Mix, we do not employ input-level mixup. } \label{ablation:randomchannelmix}
\end{table}

\begin{figure}[t]
    \centering
    \vspace{-0.2cm}
    \includegraphics[width=0.8\linewidth]{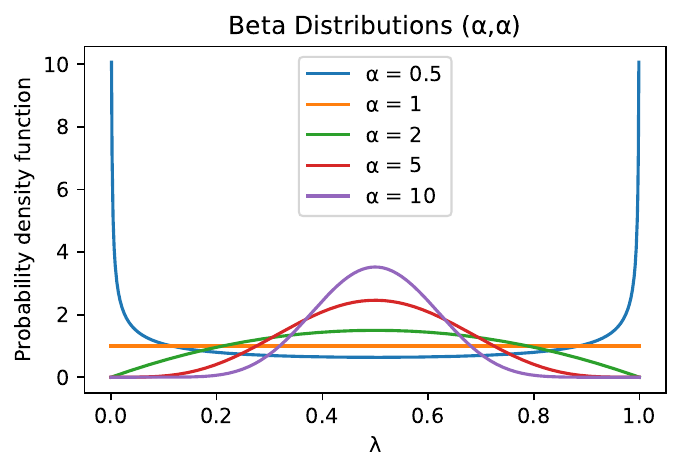}
    \vspace{-0.2cm}
    \caption{Beta distributions with alpha $\alpha$. } \label{fig:beta}
\end{figure}

\subsection{Hyper-parameter Setting: $\alpha$}
\paragraph{Relation between $\lambda$ and $\alpha$} We investigate the impact of the hyperparameter $\alpha$, which determines the distribution to sample the mixing ratio $\lambda \sim Beta(\alpha,\alpha)$.
The mixing ratio $\lambda$ determines the extent to which images are mixed together during training.
As in Figure~\ref{fig:beta}, a larger $\alpha$ draws $\lambda$ closer to a value of 0.5. 
This suggests an equal 50/50 mix of the source and target, posing a challenging sample. Conversely, a smaller $\alpha$ results in easier samples. 
In summary, $\alpha$ determines the intensity with which the mixup is applied.

\paragraph{Hyperparameter search} 
Hyperparameter $\alpha$ is searched for [0.1, 1.0, 2.0, 5.0, 10.0].
We use PreActResNet-18 for CIFAR-100 and Tiny-ImageNet, and use ResNet-18 for CUB, CARS, and Aircraft dataset, following training protocol in Table 1,2,3 of main paper.
Table~\ref{tab:alpha} show that high alpha ($\alpha$ = 10) is suitable for all datasets except CIFAR-100 ($\alpha$ = 0.5). However, this isn't due to the sensitivity of our method's hyper-parameter but rather a common trend noticed when working with CIFAR at a 32x32 resolution. For instance, in the official implementations of CutMix and SaliencyMix, the mixing probability is set to 0.5 exclusively for the CIFAR-100 and 1.0 for others, intending to perform a lower level of regularization for CIFAR. 
Given that a smaller alpha in our approach indicates easier samples, it's understandable that CIFAR-100 would require a different alpha. 

\begin{table}[t]
\centering
    \small
    % \resizebox{\linewidth}{!}{
    \begin{tabular}{rccccc}
        \toprule
        & \multicolumn{2}{c}{PreActResNet18} & \multicolumn{3}{c}{ResNet18}\\
         \cmidrule(lr){2-3} \cmidrule(lr){4-6}
        % & \textbf{CIFAR} & \textbf{TI} & \textbf{CUB} & \textbf{Cars} & \textbf{Aircraft} \\
        \multicolumn{1}{c}{\textbf{$\alpha$}} & CIFAR & TI & CUB & Cars & Airs \\
        \midrule
        {0.5}  & 82.21 & 68.55 & 69.99 & 87.32 & 81.64 \\
        {1.0}  & \textbf{82.24} & 67.60 & 71.13 & 87.44 & 81.85 \\
        {2.0}  & 81.89 & 68.37 & 70.78 & 87.63 & 81.55 \\
        {5.0}  & 82.19 & 69.19 & 72.20 & 87.51 & 81.43 \\
        {10.0} & 81.49 & \textbf{69.27} & \textbf{72.44} & \textbf{87.71} & \textbf{82.21} \\
        \bottomrule
    \end{tabular}
    \caption{Hyper-parameter search $\alpha$, top-1 accuracy rate (\%, $\uparrow$) on diverse datasets. Mixing ratio $\lambda$ for Catch-up Mix is sampled from Beta distribution $\lambda \sim Beta(\alpha,\alpha)$. }
    \label{tab:alpha}
    % }
\end{table}

\begin{table}[t]
\centering
\begin{tabular}{lc}
\toprule
Layer Set & Top-1 Acc (\%) \\
\midrule
Vanilla & 77.01 \\
\midrule
\multicolumn{2}{l}{\textit{\textbf{Catch-up Mix}}} \\
\(K=\{0,1,2,3,4\}\)          & 82.36 \\
\(K=\{0,1,2,3,5\}\)          & 81.82 \\
\(K=\{0,1,2,4,5\}\)          & 81.35 \\
\(K=\{0,1,3,4,5\}\)          & 82.00 \\
\(K=\{0,2,3,4,5\}\)          & 81.72 \\
\(K=\{0,1,2,3,4,5\}\) (Ours) & 82.41 \\
\bottomrule
\end{tabular}
\caption{Ablation study of composing different mixup layer sets using PreActResNet-18. We follow the settings used in Table~\ref{ablation:randomchannelmix}.}
\label{ablation:layer_set}
\vspace{-0.2cm}
\end{table}

\subsection{Composing Different Mixup Layer Sets}
In this paper, we use our method with the full mixup layer set over all the experiments. Additionally, we explore the effects of composing different layer sets to conduct Catch-up Mix. We perform ablation experiments on CIFAR-100, which is utilized as a generalization performance measurement. We subtract single component from the full mixup layer set $K=\{0,1,2,3,4,5\}$, and compare with each other, original Catch-up Mix and Vanilla method in Table~\ref{ablation:layer_set}. Note that we sample $k \sim K$ with the uniformly increased probability when we subtract the component $k$. With the full mixup layer set, our method achieves better performance than other layer sets.

\subsection{Catch-up Mix in Multi-stage Architecture}
We further tested our method on various architectures beyond the 4-stage ones, like ResNet variants, as shown in Table 1, 2 and 6 of the main paper. Additionally, we conducted experiments on MobileNetV2, and VGG-11 using CUB200. Both architectures have more than 4-stages which are commonly used in deep learning field. We follow the experimental settings described in Table 3 of main paper. MobileNetV2 consists of 8 stages including stem layer, and VGG-11 consists of 8 different convolution layers. Following our approach, we conduct mixup in the mixup layer which is uniformly sampled from the full layer set, and use the same hyper-parameter $\alpha=10$. From the result in Table~\ref{ablation:multi_stage}, Catch-up Mix performs also well in the multi-stage architecture without any manipulation.

\begin{table}[t]
\small
\centering
\begin{tabular}{lcc}
    \toprule
    Architecture & Vanilla & +Catch-up Mix \\
    \midrule
    MobileNetV2   & 65.53 & 71.26 (+5.73) \\
    VGG-11        & 61.01 & 67.73 (+6.72) \\
    \bottomrule
\end{tabular}
\caption{Top-1 accuracy rate (\%) on CUB200 using multi-stage architecture MobileNetV2 and VGG-11.}
\label{ablation:multi_stage}
\end{table}

\subsection{Comparison with Regularization Methods}

Here, we compare our approach with regularization methods other than mixup. 
Previous studies~\cite{li2021feature_moex,yun2019cutmix} have used PyramidNet-200 to evaluate the effectiveness of different regularization strategies, including stochastic depth, label smoothing, cutout, and dropblock.
We have replicated the Catch-up Mix in the same setting. 
Both results show that Catch-up Mix outperforms other non-mixup regularization methods.

\begin{table}[t]
\centering
\begin{tabular}{lc}
    \toprule
    \multirow{2}{*}{\textbf{Method}} & \textbf{Top-1 } \\
    & \textbf{Acc (\%)} \\
    \midrule
    Vanilla & 83.55 \\
    StochDepth & 84.14 \\
    Label Smoothing & 83.27 \\
    Cutout & 83.47 \\
    DropBlock & 84.27 \\
    ShakeDrop & 83.86 \\
    Mixup & 84.37 \\
    ManifoldMixup & 83.86 \\
    CutMix & \underline{85.53} \\
    MoEx & 84.98 \\
    \midrule
    \textbf{Catch-up Mix} & \textbf{85.89} \\
    \bottomrule
\end{tabular}
\caption{Top-1 accuracy rate (\%) on CIFAR-100 using PyramidNet-200 (alpha=240). All the other results are from \cite{li2021feature_moex} and \cite{yun2019cutmix}.}
\end{table}

\end{document}